\documentclass{article}

    \PassOptionsToPackage{numbers, compress}{natbib}

\usepackage[main,final]{neurips_2026}

\usepackage{microtype}
\usepackage{graphicx}
\makeatletter
\@ifundefined{Gread@transgrouptrue}{}{%
  \let\Gread@transgrouptrue\Gread@transgroupfalse
}
\makeatother
\usepackage{subcaption}
\usepackage{booktabs} %
\usepackage{xcolor}
\usepackage{multirow}   %
\usepackage{siunitx}    %
\usepackage{tabularx}   %
\usepackage{array}      %
\usepackage{caption}    %
\usepackage[table]{xcolor}
\usepackage{makecell}
\usepackage{hyperref}
\usepackage{float}
\definecolor{lightgreen}{RGB}{220,245,220}

\renewcommand{\answerYes}[1][]{[Yes]#1}
\renewcommand{\answerNo}[1][]{[No]#1}

\usepackage{amsmath}
\usepackage{amssymb}
\usepackage{mathtools}
\usepackage{amsthm}
\usepackage{wrapfig}
\usepackage{needspace}
\usepackage{soul}
\usepackage{adjustbox}
\usepackage{lipsum}

\usepackage[capitalize,noabbrev]{cleveref}

\DeclareRobustCommand{\contribution}[1]{#1}

\newtheorem{definition}{Definition}
\newtheorem{theorem}{Theorem}
\newtheorem{proposition}{Proposition}

\AddToHook{env/definition/begin}{\crefalias{definition}{definition}}
\AddToHook{env/theorem/begin}{\crefalias{theorem}{theorem}}
\AddToHook{env/proposition/begin}{\crefalias{proposition}{proposition}}
\AddToHook{env/remark/begin}{\crefalias{remark}{remark}}

\title{On the Information Loss of Multi-Token Prediction: Origin and Solution}

\author{%
  \textbf{Tangyu Jiang}$^{1}$ \quad
  \textbf{Zhanke Zhou}$^{1}$ \quad
  \textbf{Haodi Wang}$^{2}$ \\
  \textbf{Yiu-ming Cheung}$^{1}$ \quad
  \textbf{Xiaohua Jia}$^{2}$ \quad
  \textbf{Bo Han}$^{1}$\thanks{{\fontsize{9}{10}\selectfont Correspondence to Bo Han (bhanml@comp.hkbu.edu.hk).}} \\
  $^{1}$Department of Computer Science, Hong Kong Baptist University \\
  $^{2}$Department of Computer Science, City University of Hong Kong
}

\hypersetup{%
  pdfborder={0 0 0},
  pdftitle={On the Information Loss of Multi-Token Prediction: Origin and Solution},
  pdfauthor={Tangyu Jiang, Zhanke Zhou, Haodi Wang, Yiu-ming Cheung, Xiaohua Jia, Bo Han}
}

\begin{document}

\setcounter{footnote}{1}
\maketitle
\setcounter{footnote}{1}

\begin{abstract}
Multi-token prediction (MTP) accelerates autoregressive decoding through block-wise prediction, but remains vulnerable to error accumulation across sequentially generated blocks.
To investigate this limitation, we analyze the distinction between block-wise and sequence-level cumulative information loss (CIL). Specifically, when the model generates using its own previous outputs without access to reference answers, the input-conditional gap equals the cross-block dependence of the reference targets. To improve prediction when these generated prefixes are imperfect, we propose \textit{Dual-MTP}, which supervises shared prediction heads under full and randomly masked contexts. Together, complete-prefix supervision and context perturbations provide complementary training signals while retaining the MTP architecture. To characterize this training objective, we use the normalized dual loss to bound the divergence between the model's predictive distributions under full and masked contexts. Experiments across five backbones and six tasks spanning mathematical reasoning and GLUE understanding show that Dual-MTP achieves the highest average task score on four of five backbones, with gains of up to $2.46$ points over MTP. On mathematical reasoning tasks, Dual-MTP also achieves $1.4\times$--$1.9\times$ the decoding throughput of the NTP reference. Together, these results show that dual-context supervision can improve prediction accuracy while retaining efficient multi-token decoding.
Code is available at \url{https://github.com/tmlr-group/dual-mtp}.
\end{abstract}

\section{Introduction}

Large Language Models (LLMs) \cite{yang2025qwen3, jaech2024openai, guo2025deepseek} have achieved remarkable success across a wide range of tasks \cite{yi2025survey,wang2023review,zhao2023more,feng2026rewardflow}.
The standard training paradigm for these models is Next-Token Prediction (NTP), in which each token is predicted conditioned on the preceding tokens available within the model\textquotesingle s context window \cite{team2024qwen2, he2025law, minhturning}.
While effective, NTP provides only a single-token supervision signal at each position, which may limit the richness of the learned representations, particularly for tasks that require multi-step reasoning \cite{zhou2026landscape}.

Multi-Token Prediction (MTP) has recently emerged as a compelling alternative to the NTP objective~\cite{gloeckle2024better, liu2024deepseek, xiao2026mimo}. Rather than predicting the next token at each position, MTP partitions the target sequence into blocks of multiple future tokens and trains the model to forecast an entire block at once, providing denser supervision per position. Concretely, MTP employs a shared backbone with multiple draft heads, where each head predicts a different token from the same hidden representation. This intra-block multi-target supervision requires a single representation to jointly support all correlated prediction targets, encouraging the model to capture richer structures among tokens within each block. Through this mechanism, MTP learns more informative representations for multi-step prediction and has demonstrated strong empirical performance at scale.

\begin{figure}[t]
    \centering
    \begin{minipage}[t]{0.6\linewidth}
        \centering
        \includegraphics[width=\linewidth]{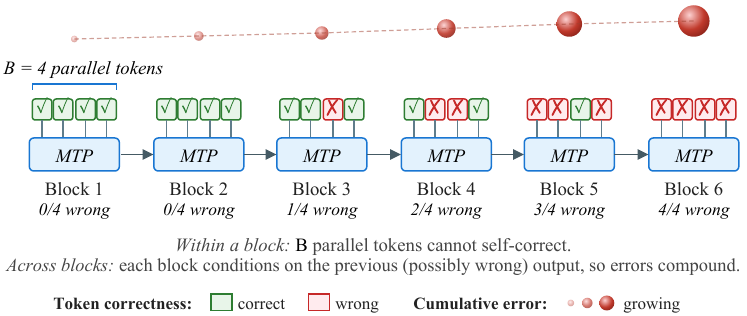}
    \end{minipage}
    \hfill
    \begin{minipage}[t]{0.39\linewidth}
        \centering
        \begin{minipage}[t]{\linewidth}
            \centering
            \includegraphics[width=\linewidth]{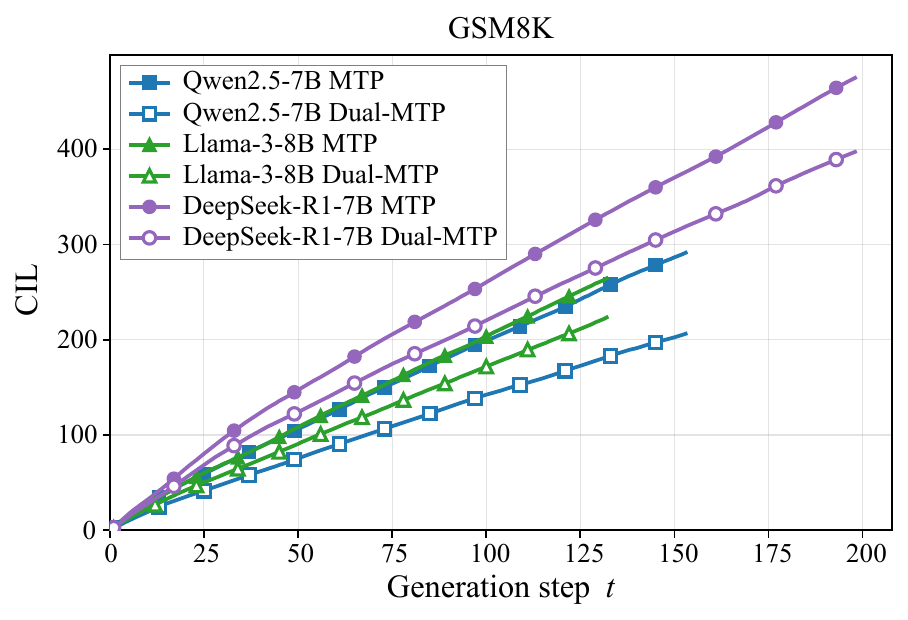}
        \end{minipage}
    \end{minipage}

    \caption{
    \textbf{Cumulative error in block-wise multi-token prediction.}
    \textbf{Left:} Example of snowball error, where early block-level mistakes propagate and amplify across subsequent predictions.
    \textbf{Right:} Visualization of the empirical cumulative predictive-entropy proxy across different models.
    }
    \label{fig:snowball}
    \vspace{-21pt}
\end{figure}

In this paper, we revisit MTP through an information-theoretic lens.
Under teacher forcing, each block-level representation is optimized with access to a clean ground-truth prefix.
At inference time, however, this prefix is replaced by the model’s predictions.
Parallel offset heads propose tokens from a shared representation, and accepted tokens become the context for subsequently generated tokens.
Consequently, even a small error in an early block can perturb the conditioning distribution that the representation was trained to expect.
If predictions are sensitive to errors in the generated prefix, these perturbations can propagate through subsequently generated tokens and accumulate across successive blocks, producing a \emph{snowball error} that degrades generation quality~\cite{ganrethinking,tian2025multi} (see \cref{fig:snowball}).
This motivates training the prediction heads to remain effective under incomplete or imperfect historical context~\cite{zhou2025from}.

We propose \textbf{Dual-MTP}, which jointly supervises the same targets under full and masked historical contexts using shared parameters. This objective combines learning from complete prefixes with supervision under context perturbations. We use the normalized dual loss to bound the expected Jensen--Shannon divergence between the predictions made by the same head under full and masked contexts. Across five backbones and six benchmarks, Dual-MTP achieves the highest average task score on four backbones, with gains of up to $2.46$ points over MTP.

In summary, we make the following contributions:
\begin{itemize}
    \item \contribution{We express the CIL gap as conditional dependence among reference-target blocks when generation uses model outputs without reference answers. We further derive an upper bound on the divergence between full- and masked-context predictive distributions, linking the dual training objective to prediction consistency (\cref{sec:theory,sec:Dual-MTP}).}

    \item \contribution{We propose Dual-MTP, which trains shared heads to predict the same targets under full and masked contexts. In this way, it combines complete-prefix supervision with robustness to context perturbations without adding inference-time parameters (\cref{sec:Dual-MTP}).}

    \item \contribution{We evaluate Dual-MTP across models and benchmarks, showing average-score gains on four of five backbones. Together, these experiments characterize the benefits and limitations of dual-context supervision across tasks and training configurations} (\cref{sec:experiments}).
\end{itemize}

\section{Related Work}

\subsection{Multi-token Prediction}

MTP enables a single model to predict multiple future tokens within one decoding step while remaining compatible with autoregressive verification. 
Prior work has explored two main paradigms to improve the performance of MTP: post-training optimization and native pre-training. 
Post-training methods augment pretrained LLMs with auxiliary draft heads and fine-tune only these components to extract latent future-token information \cite{ cai2024medusa, liu2024deepseek}, building on early blockwise parallel decoding \cite{stern2018blockwise}. 
Recent extensions adapt this paradigm for speculative decoding, including PaSS for parallel multi-token drafting \cite{monea2023pass}, and leveraging non-adjacent context for longer-horizon prediction \cite{liu2025lmtp}. 
Complementary to these, native pre-training approaches incorporate multi-token objectives directly into training, as in ProphetNet \cite{qi2020prophetnet}, Meta MTP \cite{gloeckle2024better}, DeepSeek-V3 \cite{liu2024deepseek}, and MiMo \cite{xiao2026mimo}. Compared to generic speculative decoding, MTP offers structured parallelism without external draft models.
Dual-MTP follows the post-training paradigm: it adapts a pretrained backbone with LoRA and jointly trains the added prediction heads.
However, existing pre-training approaches for MTP are still grounded in local conditional likelihood objectives, focusing on improving per-step or intra-block prediction accuracy. 
Such objectives do not explicitly capture how errors propagate across blocks, leaving the global dependency structure under-constrained.
These observations motivate training for robustness to imperfect historical context. Our dual-context objective supervises the prediction heads under both full and perturbed prefixes.

\subsection{Parallel Decoding}

The sequential nature of NTP has motivated extensive research on parallel decoding to improve inference efficiency. 
Existing approaches typically aim to amortize decoding cost by generating or verifying multiple tokens per step, often following a draft-and-verify paradigm \cite{cai2024medusa, chen2023accelerating, chen2024cascade, geng2021learning, gu2019levenshtein}. 
Representative methods include speculative sampling \cite{leviathan2023fast, chen2023accelerating, pmlr-v235-li24bt, li2024eagle, li2025eagle}, tree-based verification \cite{wang2025opt, xiong2025dyspec, chen2024sequoia, brown2024dynamic}, which differ in how draft tokens are produced and validated, and pipelined-based approaches \cite{yangpredictive, butler2024pipeinfer, monea2023pass, yi2024generation}. 
While these techniques can yield notable speedups under favorable conditions, they often require carefully calibrated auxiliary models, and the associated verification overhead can limit their robustness. 
As a result, practical efficiency gains remain highly sensitive to model behavior and task characteristics \cite{zhou2025alphaapollo,zhou2026alphadiana,zhou2026codapo}.

\section{Preliminaries}
 \label{sec:preliminaries}
\textbf{Next-Token Prediction (NTP).} 
An autoregressive language model factorizes the joint distribution over an output sequence $\mathbf{x} = (x_1, \ldots, x_L)$ as
\begin{equation}
p_\theta(\mathbf{x}) = \prod_{l=1}^{L} p_\theta(x_l \mid x_{<l}).
\end{equation}
Here $x_{<l}=(x_1,\ldots,x_{l-1})$ denotes the preceding tokens available within the context window. Training uses reference prefixes; inference uses the model\textquotesingle s own preceding outputs. The training objective is maximum likelihood estimation, which minimizes the negative log-likelihood
\begin{equation}
\mathcal{L}_{\text{NTP}}(\theta) 
= - \mathbb{E}_{\mathbf{x} \sim \mathcal{D}} 
\sum_{l=1}^{L} \log p_\theta(x_l \mid x_{<l}),
\end{equation}
where $\mathcal{D}$ denotes the training corpus. Since each position is supervised by a single target token, NTP provides one-token training signal per step and decodes tokens sequentially at inference.

\textbf{Multi-Token Prediction (MTP).} MTP extends NTP by partitioning the target sequence into non-overlapping blocks of $B$ consecutive tokens and jointly predicting each block from its prefix, providing denser supervision per position. The training objective is the block-wise negative log-likelihood
\begin{equation}
\mathcal{L}_{\text{MTP}}(\theta)
= - \mathbb{E}_{\mathbf{x} \sim \mathcal{D}}\sum_{s}
\log p_\theta(x_{l_s:l_s+B-1} \mid x_{<l_s}),
\end{equation}
which is typically realized by $D$ parallel prediction heads sharing a backbone. At inference time, this design enables parallel decoding and reduces the number of sequential steps by up to a factor of $B$, but tokens within a block are not conditioned on verified realizations of earlier tokens in the same block. Teacher forcing supplies reference prefixes during training. At inference, parallel heads do not consume one another\textquotesingle s proposed tokens within a block, but committed tokens form the \mbox{context for later blocks}.

\begingroup{}
\textbf{Reference Answers and Model Outputs.}
Let $X$ denote a test input and $T=(t_1,\ldots,t_N)$ its reference answer, sampled together from $p_{\mathrm{data}}$. With the trained parameters $\theta$ fixed, the model generates $R_N=(r_1,\ldots,r_N)$ from $X$ and its own preceding outputs, without access to $T$. Any decoding randomness is independent of $T$ given $X$. Thus, $T$ is supplied by the data, whereas $R_N$ is produced by the model. For theoretical analysis, we represent both sequences using fixed blocks of token positions, with a padding convention for positions beyond the end of each sequence. The verifier may accept a different number of tokens at each decoding step, so these fixed blocks need not match the accepted token groups.
\endgroup{}

\textbf{Cumulative Information Loss.}
Following \cite{ganrethinking}, the per-step information loss at step $l$ is defined as $\mathrm{InfoLoss}(r_l) = H(t_l) - I(t_l; r_l) = H(t_l \mid r_l)$, where $t_l$ and $r_l$ denote the target and predicted token random variables, and $r_l$ is conditioned on the partial response $R_{l-1}=(r_1,\dots,r_{l-1})$. This quantity measures residual target uncertainty given the predicted token. 
Then the Cumulative Information Loss (CIL) up to the $l$-th step is defined as 
\begin{equation}\label{eq:cum_ntp}
H_{<l}(T \mid R) = \sum_{i=1}^{l-1} H(t_i \mid r_i),
\end{equation}
where $H(t_i \mid r_i)$ denotes the entropy of the target token $t_i$ given the predicted token $r_i$ at the $i$-th step.
CIL quantifies the accumulated uncertainty over multiple steps during the generation. 
A larger CIL indicates greater accumulated uncertainty along the generation trajectory.

\section{Information-Theoretic Analysis of MTP}
\label{sec:theory}
 
We relate the block-sequence CIL gap to conditional dependence among reference-target blocks when generation uses the model\textquotesingle s own outputs without reference answers.

\subsection{Block Decomposition and the Information Gap}
 
Following the CIL in Equation \eqref{eq:cum_ntp}, we analyze MTP analogously in the information-theoretical perspective.
We first define the CIL within each block. 
For MTP, where the model predicts multiple tokens with block size $B$, the CIL within the block is defined below.
\begin{definition}[CIL within Blocks]\label{def:cum_mtp}
The block-level information loss for a block of size $B$ at the $l$-th step is defined as the information loss over the entire block:
\begin{equation}
 \mathrm{InfoLoss}(R_{l}^{B}) = H(T_{l}^{B}) - I(T_{l}^{B}; R_{l}^{B}) = H(T_{l}^{B} \mid R_{l}^{B})
\end{equation}
where $T_l^B$ denotes the block of reference tokens starting at position $l$, and $R_l^B$ denotes the corresponding block of tokens predicted by the model. 
\end{definition}

Definition~\ref{def:cum_mtp} measures the uncertainty in the reference tokens given the tokens predicted by the model for that block.
\begingroup{}
For the same target block $T_l^B=(t_l,\ldots,t_{l+B-1})$ and predicted-token block $R_l^B=(r_l,\ldots,r_{l+B-1})$, the chain rule gives
\begin{equation}\label{eq:block-token-cil}
 H(T_l^B\mid R_l^B)
 =\sum_{i=l}^{l+B-1}H(t_i\mid t_{l:i-1},R_l^B)
 \le\sum_{i=l}^{l+B-1}H(t_i\mid r_i).
\end{equation}
Each $r_i$ is contained in $R_l^B$. Thus, for the same target block and predicted-token block, the sum of token-wise conditional entropies upper-bounds block-level CIL.
\endgroup{}

However, such block-wise modeling also introduces a structural discrepancy, as the sequence is decomposed into multiple blocks. While the information loss within each block characterizes the local prediction quality, it does not account for the potential loss induced by this decomposition.
To quantify this effect, we define $\Delta$ as the difference between the sum of per-block information loss and the information loss when treating the entire sequence as a single block.
Suppose a sequence is partitioned into $M$ non-overlapping blocks.
We propose the following definition.

\begin{definition}[Block-Sequence CIL Gap]\label{def:cum_intrablock}
	Let $X$ denote the input query, shared across all blocks. Writing $\mathrm{InfoLoss}(R\mid X):=H(T\mid R, X)$ for the input-conditional version of the loss in \cref{def:cum_mtp},
	the block-sequence CIL gap for a sequence of $M$ blocks is defined as:
	\begin{equation}\label{eq:delta-mi}
		\Delta := \sum_{m=1}^M \mathrm{InfoLoss}(R_m^B\mid X) - \mathrm{InfoLoss}(R_N\mid X).
	\end{equation}
\end{definition}
 
The second term on the right-hand side follows from the chain rule of conditional entropy, which is
 \begin{equation}
 H(T_{N}\mid R_{N},X)
 = \sum_{m=1}^M H(T^{B}_{m}\mid T^{B}_{<m},R_{N},X),
 \end{equation}
where $T_N=(T^B_1,\dots,T^B_M)$ and $R_N=(R^B_1,\dots,R^B_M)$. Thus,
 \begin{equation}
 \Delta= \sum_{m=1}^M
 I(T^B_{m};T^B_{<m},R^B_{\neq m}\mid R^B_{m},X).
 \end{equation}
\begingroup{}
\noindent\textbf{Remark 1.} For each reference block $T_m^B$, compare conditioning on input $X$ and its prediction $R_m^B$ with also including preceding references $T_{<m}^B$ and other predictions $R_{\ne m}^B$. The gap sums these uncertainty reductions.
\endgroup{}

\subsection{Main Result}

\begingroup{}
\begin{theorem}[Reference-target dependence]\label{thm:main}
Let $(X,T)\sim p_{\mathrm{data}}$ and fix the trained parameters $\theta$ and the block partition. When generation uses only $X$ and the model\textquotesingle s own outputs, with decoding randomness independent of $T$ given $X$, $R_N\perp T\mid X$, and
\begin{equation}\label{eq:thm-main-bound}
\Delta=\sum_{m=1}^{M} I(T_m^B;T_{<m}^B\mid X).
\end{equation}
\end{theorem}
\textit{Proof sketch.} Both the draft heads and the verifier generate from $X$ and the preceding model outputs. Given $X$, the predictions are independent of the reference answer, so the two terms in $\Delta$ simplify to $\sum_m H(T_m^B\mid X)$ and $H(T\mid X)$. The chain rule expands the second term one block at a time. Subtracting gives the information that earlier reference blocks provide about each block given $X$. The full proof is in \cref{app:proof-thm1}.

\noindent\textbf{Remark 2.} For a fixed reference distribution and block partition, $\Delta$ is determined by the conditional dependence among reference-target blocks and is invariant to $\theta$. In \cref{subsec:dual-training}, we use the training loss to bound the divergence between the predictions made under full and masked contexts.

\endgroup{}

\section{Dual-Perspective MTP}
\label{sec:Dual-MTP}

\textbf{Dual-MTP} trains shared prediction heads to predict the same targets from full and masked historical contexts. The dual objective combines complete-prefix supervision with context perturbations, encouraging the heads to remain predictive when parts of the historical context are unavailable.

\begin{figure}[t]
    \centering
    \includegraphics[width=\textwidth]{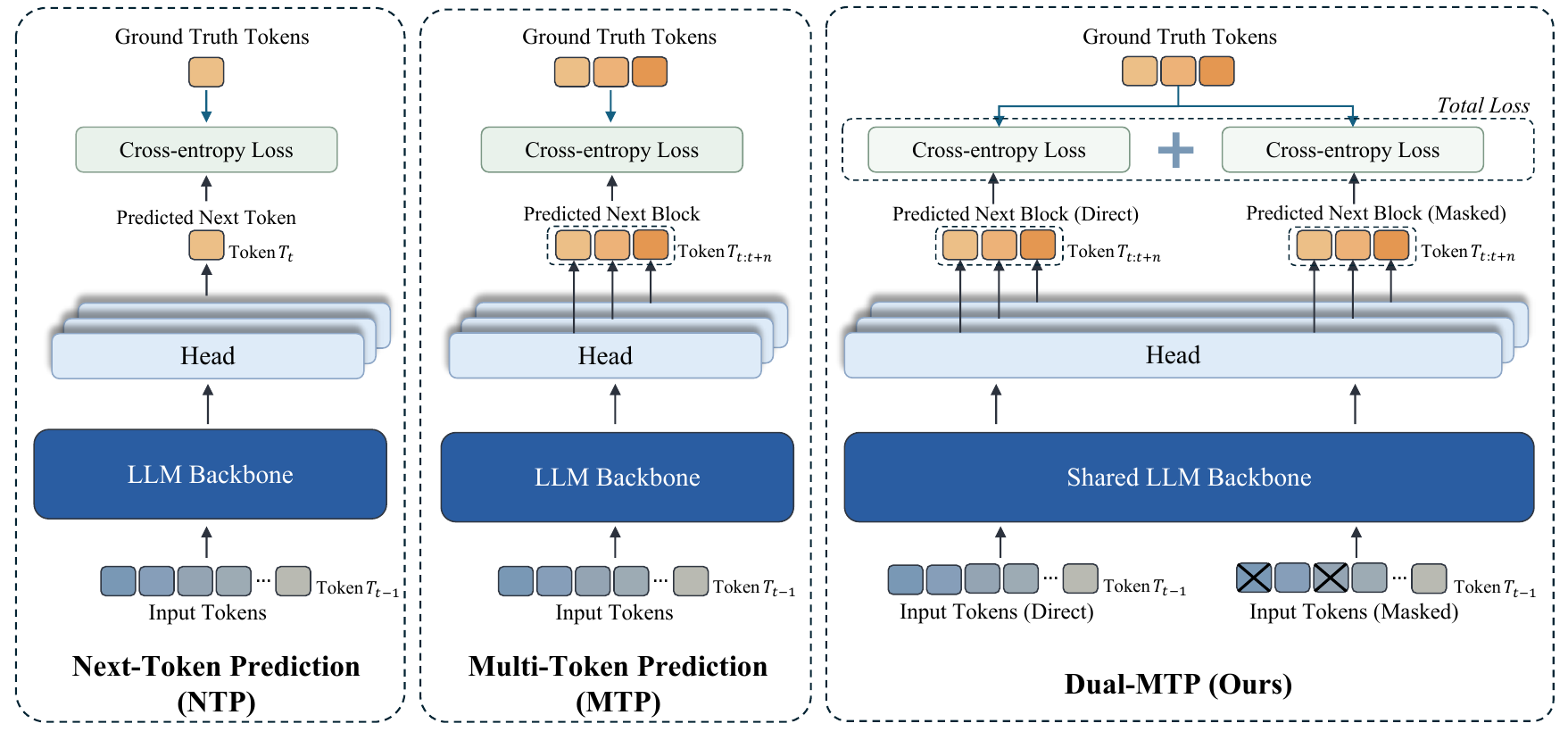}
    \caption{\textbf{Overview of NTP, MTP, and Dual-MTP.} \textit{Left:} NTP predicts one next token. \textit{Middle:} MTP predicts a token block via multiple heads on a shared backbone. \textit{Right:} Dual-MTP uses full and masked training views that share the backbone and prediction heads.
    }
    \label{fig:Dual-MTP-overview}
    	\vspace{-10pt}
\end{figure}

Dual-MTP keeps the MTP architecture, a shared autoregressive backbone with parallel multiple draft heads, and operates during training.
Different from the MTP, each head in Dual-MTP is supervised under both the full historical context and a stochastically masked version of it, encouraging block-level representations that encode compact, non-redundant information rather than relying on full prefix access. 
An overview of our method and the comparisons are shown in \cref{fig:Dual-MTP-overview}.

\Needspace{8\baselineskip}
\subsection{Dual-Perspective Training}
\label{subsec:dual-training}

\textbf{Token Stochastic Masking.} To realize the masked-context branch, we apply \emph{token stochastic masking} during training: at each step, a single binary mask is sampled over token positions with rate $\gamma$ and shared across all draft heads, blocking attention to the masked positions while leaving the standard autoregressive structure intact at inference. The masking rate $\gamma$ follows an annealing schedule with temperature-controlled perturbations to encourage early-stage exploration. Full details of the schedule and the attention-level masking are provided in Appendix \ref{app:token-masking}.

\textbf{Dual-Perspective Training Objective.} To stabilize training under stochastic masking, we combine supervision from both full-context and masked-context views \cite{zhang2026co}. The \emph{full-context loss} $\mathcal{L}_{\text{full}}=-\sum\log p(\mathbf{x}_{t:t+B-1}\mid\mathbf{x}_{< t})$ uses the complete historical context $\mathbf{x}_{< t}$, whereas the \emph{masked-context loss} $\mathcal{L}_{\text{masked}}=-\sum\log p(\mathbf{x}_{t:t+B-1}\mid\tilde{\mathbf{x}}_{< t})$ replaces $\mathbf{x}_{< t}$ with its stochastically masked version $\tilde{\mathbf{x}}_{< t}$. The overall objective is
\begin{equation}
\mathcal{L} = \mathcal{L}_{\text{full}} + \lambda(s)\,\mathcal{L}_{\text{masked}},
\end{equation}
where $\lambda(s)$ is a warm-up coefficient. The two losses play complementary roles: $\mathcal{L}_{\text{full}}$ enforces \emph{sufficiency}, ensuring block-level representations remain predictive given the complete history, while $\mathcal{L}_{\text{masked}}$ acts as a regularizer that discourages over-reliance on specific past tokens.
Both losses update the shared backbone and prediction heads. The full-context branch retains supervision on the uncorrupted prefix as the masked branch learns from incomplete histories. The masked-only ablation in \cref{subsec:ablation} illustrates the importance of this anchor for task accuracy. Normalizing both losses over the same supervised positions allows us to analyze prediction consistency across context views.

\textbf{Dual Loss and Predictive Consistency.}
For each prediction head and each token position it supervises, let $q_f$ and $q_m$ denote its predictive distributions for the same target token $Y$ under full and masked contexts, respectively. Here $\overline{\mathcal L}_{\mathrm{full}}$ and $\overline{\mathcal L}_{\mathrm{masked}}$ are normalized averages of the two losses over the same data, masks, heads, and positions, with natural logarithms.
\begin{theorem}[Dual loss bounds Jensen--Shannon divergence]\label{thm:view-mi}
At any training step $s$ with $\lambda(s)>0$, define the normalized weighted loss
\[
L_{\lambda(s)}
=\overline{\mathcal L}_{\mathrm{full}}
+\lambda(s)\,\overline{\mathcal L}_{\mathrm{masked}}.
\]
Then the expected Jensen--Shannon divergence between the two predictive distributions satisfies
\begin{equation}\label{eq:view-mi-bound}
\mathbb E[\operatorname{JS}(q_f,q_m)]
\le (\ln 2)\left[1-\exp\!\left(-\frac{L_{\lambda(s)}}{\min\{1,\lambda(s)\}}\right)\right].
\end{equation}
\end{theorem}
\textit{Proof sketch.} For a target token $y$, the overlap between the two distributions bounds their JS divergence by $(\ln 2)(1-q_f(y)q_m(y))$. The product equals $e^{-\ell}$, where $\ell$ is the sum of the two token losses. Averaging and applying Jensen's inequality gives the stated bound because $\mathbb E[\ell]\le L_{\lambda(s)}/\min\{1,\lambda(s)\}$. The full proof is in \cref{app:view-mi}.

\noindent\textbf{Remark 3.} At $\lambda(s)=1$, the bound depends on the unweighted sum of the two losses. As $L_{\lambda(s)}/\min\{1,\lambda(s)\}\to0$, the expected divergence between the two predictions vanishes. Full-context supervision anchors prediction on the complete prefix, while masked-context supervision trains the same heads to predict from partial history.

\subsection{Inference}
\label{subsec:inference}
 
\begingroup{}
Dual-MTP uses draft-and-verify inference~\cite{stern2018blockwise}. The primary head and $D$ parallel draft heads propose an ordered block of $D+1$ tokens. The verifier checks the candidates conditioned on their preceding prefix and commits the longest accepted prefix. The verification parameter $K_v$ specifies the top-$K_v$ candidate set. At $K_v=1$, multiple consecutive draft tokens can match the verifier's top-1 choices and be accepted in one iteration, reducing latency. With greedy primary-token selection, greedy fallback, and consistent tie-breaking, exact verification preserves the fixed verifier's greedy output. For $K_v>1$, the verifier can accept additional candidates within its top-$K_v$ set. The accepted tokens influence subsequent prefixes, making $K_v$ a control on the accuracy--speed trade-off. We evaluate this trade-off in \cref{app:verification-parameter}.
\endgroup{}

\begin{table*}[t]                                            
	\centering
	\caption{Main results on task performance. All results are reported as mean (\%) $\pm$ std. Avg. is the arithmetic mean over six tasks; SD is over repeated decoding runs.}  
	\label{tab:main_results}       
	\resizebox{\linewidth}{!}{       
\begin{tabular}{c|l|cc|cccc|c}                                   
\toprule                                                                  
Model & Setting & Math500 & GSM8K & CoLA & MRPC & RTE & WNLI & \textbf{Avg. $\uparrow$} \\                                               
\midrule                                                                  
\multirow{4}{*}{\makecell[c]{Llama-3.2-1B\\-Instruct}}
& Base & 13.47{\scriptsize $\pm$1.36} & 30.78{\scriptsize $\pm$2.29} & 78.88{\scriptsize $\pm$1.65} & 77.21{\scriptsize $\pm$3.19} & 66.43{\scriptsize $\pm$4.72} & 55.40{\scriptsize $\pm$5.86} & 53.69 \\                                                 
& MTP & \textbf{14.40{\scriptsize $\pm$0.35}} & 31.51{\scriptsize $\pm$1.75} & \textbf{79.51{\scriptsize $\pm$1.69}} & 79.98{\scriptsize $\pm$2.06} & 68.59{\scriptsize $\pm$4.38} & 53.52{\scriptsize $\pm$8.80} & 54.59 \\

& Medusa & 12.27{\scriptsize $\pm$1.67} & 29.52{\scriptsize $\pm$0.90} & 79.39{\scriptsize $\pm$0.63} & 77.94{\scriptsize $\pm$0.65} & \textbf{72.44{\scriptsize $\pm$0.91}} & \textbf{56.81{\scriptsize $\pm$6.95}} & 54.73 \\ 

& \cellcolor{lightgreen}Dual-MTP & \cellcolor{lightgreen}14.07{\scriptsize $\pm$2.54} & \cellcolor{lightgreen}\textbf{32.35{\scriptsize $\pm$1.26}} & \cellcolor{lightgreen}79.23{\scriptsize $\pm$1.27} & \cellcolor{lightgreen}\textbf{80.56{\scriptsize $\pm$2.87}} & \cellcolor{lightgreen}71.60{\scriptsize $\pm$1.50} & \cellcolor{lightgreen}53.05{\scriptsize $\pm$9.06} & \cellcolor{lightgreen}\textbf{55.14} \\            

\midrule                    
\multirow{4}{*}{\makecell[c]{Qwen2.5-1.5B\\-Instruct}}           

& Base & 20.16{\scriptsize $\pm$0.87} & 43.54{\scriptsize $\pm$1.30} & 81.66{\scriptsize $\pm$0.58} & 82.35{\scriptsize $\pm$1.85} & 79.06{\scriptsize $\pm$1.30} & 55.87{\scriptsize $\pm$4.95} & 60.44 \\                                                 
& MTP & \textbf{20.80{\scriptsize $\pm$1.64}} & 47.64{\scriptsize $\pm$1.54} & 82.42{\scriptsize $\pm$0.29} & 83.33{\scriptsize $\pm$0.98} & \textbf{79.78{\scriptsize $\pm$1.30}} & 55.40{\scriptsize $\pm$5.33} & 61.56 \\                                                                                      
& Medusa & 16.87{\scriptsize $\pm$1.63} & 42.99{\scriptsize $\pm$0.57} & 82.04{\scriptsize $\pm$0.49} & \textbf{83.58{\scriptsize $\pm$0.98}} & 78.82{\scriptsize $\pm$2.12} & \textbf{57.28{\scriptsize $\pm$2.15}} & 60.26 \\                                                                                        
& \cellcolor{lightgreen}Dual-MTP & \cellcolor{lightgreen}20.40{\scriptsize $\pm$0.87} & \cellcolor{lightgreen}\textbf{48.72{\scriptsize $\pm$1.38}} & \cellcolor{lightgreen}\textbf{82.71{\scriptsize $\pm$0.34}} & \cellcolor{lightgreen}83.42{\scriptsize $\pm$0.79} & \cellcolor{lightgreen}\textbf{79.78{\scriptsize $\pm$1.30}} & \cellcolor{lightgreen}55.40{\scriptsize $\pm$6.66} & \cellcolor{lightgreen}\textbf{61.74} \\                                                                                                               
\midrule  
\multirow{4}{*}{\makecell[c]{DeepSeek\\-R1-Distill\\-Qwen-7B}}                                                                                             
& Base & 26.87{\scriptsize $\pm$0.61} & 64.29{\scriptsize $\pm$1.20} & 60.56{\scriptsize $\pm$0.57} & 38.48{\scriptsize $\pm$0.20} & 53.79{\scriptsize $\pm$2.06} & 56.81{\scriptsize $\pm$1.33} & 50.13 \\                                                 
& MTP & \textbf{39.27{\scriptsize $\pm$0.66}} & 68.89{\scriptsize $\pm$1.03} & 54.84{\scriptsize $\pm$1.15} & 52.70{\scriptsize $\pm$0.53} & 57.76{\scriptsize $\pm$2.36} & 55.87{\scriptsize $\pm$1.33} & 54.89 \\                                                                                                   
& Medusa & 35.13{\scriptsize $\pm$1.45} & 65.07{\scriptsize $\pm$0.19} & 59.99{\scriptsize $\pm$0.95} & \textbf{53.02{\scriptsize $\pm$1.55}} & 46.69{\scriptsize $\pm$2.47} & 54.93{\scriptsize $\pm$1.15} & 52.47 \\                                                                                                 
& \cellcolor{lightgreen}Dual-MTP & \cellcolor{lightgreen}37.93{\scriptsize $\pm$0.90} & \cellcolor{lightgreen}\textbf{69.88{\scriptsize $\pm$0.74}} & \cellcolor{lightgreen}\textbf{70.34{\scriptsize $\pm$0.46}} & \cellcolor{lightgreen}50.25{\scriptsize $\pm$1.06} & \cellcolor{lightgreen}\textbf{59.33{\scriptsize $\pm$3.12}} & \cellcolor{lightgreen}\textbf{56.34{\scriptsize $\pm$1.15}} & \cellcolor{lightgreen}\textbf{57.34} \\                                    
\midrule                                                                  
\multirow{4}{*}{\makecell[c]{Qwen2.5-7B\\-Instruct}}                      
& Base & 26.40{\scriptsize $\pm$0.35} & 53.53{\scriptsize $\pm$0.26} & 77.47{\scriptsize $\pm$5.49} & 83.01{\scriptsize $\pm$0.92} & 88.57{\scriptsize $\pm$1.19} & 74.65{\scriptsize $\pm$1.99} & 67.27 \\                                                 
& MTP & 37.53{\scriptsize $\pm$1.37} & \textbf{72.25{\scriptsize $\pm$3.43}} & 77.34{\scriptsize $\pm$5.24} & 84.64{\scriptsize $\pm$1.17} & 89.29{\scriptsize $\pm$0.34} & 77.93{\scriptsize $\pm$3.51} & 73.17 \\                                                                                                   
& Medusa & 34.20{\scriptsize $\pm$0.72} & 69.02{\scriptsize $\pm$0.93} & \textbf{86.03{\scriptsize $\pm$0.30}} & \textbf{86.11{\scriptsize $\pm$0.42}} & 89.65{\scriptsize $\pm$0.95} & 75.59{\scriptsize $\pm$2.66} & 73.43 \\                                                                                
& \cellcolor{lightgreen}Dual-MTP & \cellcolor{lightgreen}\textbf{38.67{\scriptsize $\pm$0.90}} & \cellcolor{lightgreen}71.24{\scriptsize $\pm$1.18} & \cellcolor{lightgreen}85.33{\scriptsize $\pm$0.23} & \cellcolor{lightgreen}85.21{\scriptsize $\pm$1.01} & \cellcolor{lightgreen}\textbf{89.77{\scriptsize $\pm$0.61}} & \cellcolor{lightgreen}\textbf{80.28{\scriptsize $\pm$1.99}} & \cellcolor{lightgreen}\textbf{75.08} \\                      
\midrule                                                                  
\multirow{4}{*}{\makecell[c]{Llama-3-8B\\-Instruct}}                    
& Base & 13.33{\scriptsize $\pm$0.75} & 44.00{\scriptsize $\pm$2.18} & 85.30{\scriptsize $\pm$0.61} & 88.67{\scriptsize $\pm$0.59} & 88.09{\scriptsize $\pm$1.35} & 71.83{\scriptsize $\pm$1.99} & 65.20 \\                                                 
& MTP & \textbf{20.07{\scriptsize $\pm$0.09}} & \textbf{56.46{\scriptsize $\pm$0.58}} & 85.14{\scriptsize $\pm$0.96} & 88.14{\scriptsize $\pm$0.54} & 88.93{\scriptsize $\pm$1.36} & \textbf{73.71{\scriptsize $\pm$0.66}} & \textbf{68.74} \\                                                             
& Medusa & 17.93{\scriptsize $\pm$0.12} & 54.31{\scriptsize $\pm$0.04} & 84.69{\scriptsize $\pm$0.20} & 86.60{\scriptsize $\pm$0.81} & \textbf{89.65{\scriptsize $\pm$0.74}} & 64.79{\scriptsize $\pm$3.45} & 66.33 \\     
& \cellcolor{lightgreen}Dual-MTP & \cellcolor{lightgreen}19.27{\scriptsize $\pm$0.41} & \cellcolor{lightgreen}56.10{\scriptsize $\pm$0.86} & \cellcolor{lightgreen}\textbf{85.47{\scriptsize $\pm$0.81}} & \cellcolor{lightgreen}\textbf{88.92{\scriptsize $\pm$1.02}} & \cellcolor{lightgreen}89.41{\scriptsize $\pm$0.90} & \cellcolor{lightgreen}72.30{\scriptsize $\pm$2.39} & \cellcolor{lightgreen}68.58 \\                                

\bottomrule                                                               
\end{tabular}}
\end{table*} 

\section{Experiments}
\label{sec:experiments}

\subsection{Experimental Setup}
\label{subsec:setup}

\textbf{Tasks and Models.} We evaluate Dual-MTP on a benchmark suite spanning two task families: (i) mathematical reasoning, with MATH500~\cite{lightman2023let} and GSM8K~\cite{cobbe2021training}; and (ii) natural-language understanding from GLUE~\cite{wang2018glue}, with CoLA, MRPC, RTE, and WNLI. This selection covers both short-form classification and reasoning-heavy generation, allowing us to test whether dual-context training transfers across these generation regimes. Experiments are conducted on five backbones: Qwen2.5-1.5B-Instruct, Qwen2.5-7B-Instruct, Llama-3.2-1B-Instruct, Llama-3-8B-Instruct, and DeepSeek-R1-Distill-Qwen-7B.

\textbf{Baselines and Metrics.} We compare Dual-MTP against three baselines: (i) \textbf{Base}, a supervised fine-tuning (SFT) reference in which both the backbone and the original output head are jointly trained with the standard NTP objective on the same training data and decoded autoregressively, without any draft heads; (ii) \textbf{MTP}, the post-training blockwise draft-head variant of multi-token prediction~\cite{stern2018blockwise, cai2024medusa}, which shares the same draft-head architecture as Dual-MTP but is trained without stochastic masking; and (iii) \textbf{Medusa}~\cite{cai2024medusa}, a multi-head speculative decoding method that augments the backbone with auxiliary draft heads trained jointly with the original output head.
The three MTP methods use the same pretrained backbone family and head count, with method-specific loss weights and optimization schedules. The verification-parameter experiments use the configurations described in \cref{app:verification-parameter}. 
We report Avg@1 for math reasoning, accuracy for GLUE, except Llama-3-8B Base/MTP/Dual-MTP on MRPC, which use the mean of accuracy and F1. Math uses final-answer mathematical equivalence (Appendix~\ref{app:final-answer}).

\textbf{Implementation.} We attach $D{=}3$ draft heads to the backbone, so that the backbone's primary token together with the $D$ draft tokens form a block of size $B{=}D{+}1{=}4$, and fine-tune the backbone via LoRA jointly with the newly added heads. At inference, stochastic masking is disabled, and we run the draft-and-verify procedure of \cref{subsec:inference} with $K_v{=}5$, using at most $2048$ response tokens for math reasoning and $16$ for GLUE, excluding the prompt.

\begingroup{}
\textbf{Training Cost.}
The full and masked losses are combined within each training update. Across the evaluated backbones and corpora, one epoch of Dual-MTP takes $1.26\times$--$1.50\times$ the wall-clock time of MTP. Table~\ref{tab:training-cost} (Appendix~\ref{app:training-computation-and-cost}) reports the training costs. \Cref{app:verification-parameter} evaluates the effect of $K_v$ on mathematical reasoning and code generation.
\endgroup{}

\subsection{Main Results}
\label{subsec:main-results}

\textbf{Main Results.} \Cref{tab:main_results} compares five backbones on two math and four GLUE tasks. Dual-MTP attains the highest six-task average on four backbones, with gains of $+0.18$ to $+2.46$ points over MTP. On Llama-3-8B, its average is $68.58$ versus MTP\textquotesingle s $68.74$. It improves over Medusa on all five backbones by $+0.41$ to $+4.87$ points, and over Base by $+1.30$ to $+7.81$ points.
The largest average gain over MTP occurs on DeepSeek-R1-Distill-Qwen-7B, from $54.89$ to $57.34$. On GSM8K, Dual-MTP improves on MTP for three backbones by $0.84$--$1.09$ points, but trails it on Qwen2.5-7B and Llama-3-8B by $1.01$ and $0.35$ points. On MATH500, it improves on MTP only for Qwen2.5-7B ($38.67\%$ versus $37.53\%$). GLUE contributes substantially to the average gains on larger backbones. Dual-context supervision therefore has task- and backbone-dependent benefits.

\begin{table*}[t]
  \centering
  \caption{Decoding throughput and end-task accuracy. Acc.\ is final-answer accuracy (\%); tok/s and speedup retain the original timing protocol. Values are mean $\pm$ SD; speedup is relative to Base (NTP). Bold marks the best result among MTP, Medusa, and Dual-MTP.}
  \label{tab:decoding_speedup}
  \resizebox{\linewidth}{!}{
  \begin{tabular}{l|l|cc|cc|cc|cc|cccc}
  \toprule
  \multirow{2}{*}{Model} & \multirow{2}{*}{Task}
  & \multicolumn{2}{c|}{Base}
  & \multicolumn{2}{c|}{MTP}
  & \multicolumn{2}{c|}{Medusa}
  & \multicolumn{2}{c|}{\cellcolor{lightgreen}\textbf{Dual-MTP}}
  & \multicolumn{4}{c}{Speedup} \\
  & & Acc. & tok/s & Acc. & tok/s & Acc. & tok/s & \cellcolor{lightgreen}Acc. & \cellcolor{lightgreen}tok/s
  & Base & MTP & Medusa & \cellcolor{lightgreen}\textbf{Dual-MTP} \\
  \midrule

  Llama-3.2-1B & MATH500
  & 13.47{\scriptsize $\pm$1.36} & 34.5{\scriptsize $\pm$1.0}
  & \textbf{14.40{\scriptsize $\pm$0.35}} & 48.3{\scriptsize $\pm$1.5}
  & 12.27{\scriptsize $\pm$1.67} & \textbf{59.1{\scriptsize $\pm$0.6}}
  & \cellcolor{lightgreen}14.07{\scriptsize $\pm$2.54} & \cellcolor{lightgreen}49.3{\scriptsize $\pm$0.6}
  & 1.0$\times$ & 1.4$\times$ & \textbf{1.7$\times$} & \cellcolor{lightgreen}1.4$\times$ \\

  & GSM8K
  & 30.78{\scriptsize $\pm$2.29} & 35.9{\scriptsize $\pm$0.4}
  & 31.51{\scriptsize $\pm$1.75} & 53.2{\scriptsize $\pm$3.3}
  & 29.52{\scriptsize $\pm$0.90} & \textbf{60.3{\scriptsize $\pm$1.1}}
  & \cellcolor{lightgreen}\textbf{32.35{\scriptsize $\pm$1.26}} & \cellcolor{lightgreen}55.4{\scriptsize $\pm$2.8}
  & 1.0$\times$ & 1.5$\times$ & \textbf{1.7$\times$} & \cellcolor{lightgreen}1.5$\times$ \\

  \midrule

  Qwen2.5-1.5B & MATH500
  & 20.16{\scriptsize $\pm$0.87} & 20.8{\scriptsize $\pm$0.7}
  & \textbf{20.80{\scriptsize $\pm$1.64}} & 30.2{\scriptsize $\pm$1.5}
  & 16.87{\scriptsize $\pm$1.63} & \textbf{34.8{\scriptsize $\pm$1.9}}
  & \cellcolor{lightgreen}20.40{\scriptsize $\pm$0.87} & \cellcolor{lightgreen}31.3{\scriptsize $\pm$2.3}
  & 1.0$\times$ & 1.5$\times$ & \textbf{1.7$\times$} & \cellcolor{lightgreen}1.5$\times$ \\

  & GSM8K
  & 43.54{\scriptsize $\pm$1.30} & 20.7{\scriptsize $\pm$0.3}
  & 47.64{\scriptsize $\pm$1.54} & 33.9{\scriptsize $\pm$1.5}
  & 42.99{\scriptsize $\pm$0.57} & \textbf{38.3{\scriptsize $\pm$1.9}}
  & \cellcolor{lightgreen}\textbf{48.72{\scriptsize $\pm$1.38}} & \cellcolor{lightgreen}33.4{\scriptsize $\pm$1.6}
  & 1.0$\times$ & 1.6$\times$ & \textbf{1.9$\times$} & \cellcolor{lightgreen}1.6$\times$ \\

  \midrule

  \multirow{2}{*}{\makecell[l]{DeepSeek-R1-\\Distill-Qwen-7B}} & MATH500
  & 26.87{\scriptsize $\pm$0.61} & 19.8{\scriptsize $\pm$0.2}
  & \textbf{39.27{\scriptsize $\pm$0.66}} & 31.6{\scriptsize $\pm$1.0}
  & 35.13{\scriptsize $\pm$1.45} & \textbf{37.4{\scriptsize $\pm$1.5}}
  & \cellcolor{lightgreen}37.93{\scriptsize $\pm$0.90} & \cellcolor{lightgreen}33.1{\scriptsize $\pm$0.5}
  & 1.0$\times$ & 1.6$\times$ & \textbf{1.9$\times$} & \cellcolor{lightgreen}1.7$\times$ \\

  & GSM8K
  & 64.29{\scriptsize $\pm$1.20} & 20.0{\scriptsize $\pm$0.8}
  & 68.89{\scriptsize $\pm$1.03} & 31.8{\scriptsize $\pm$4.1}
  & 65.07{\scriptsize $\pm$0.19} & \textbf{42.5{\scriptsize $\pm$0.1}}
  & \cellcolor{lightgreen}\textbf{69.88{\scriptsize $\pm$0.74}} & \cellcolor{lightgreen}37.6{\scriptsize $\pm$1.6}
  & 1.0$\times$ & 1.6$\times$ & \textbf{2.1$\times$} & \cellcolor{lightgreen}1.9$\times$ \\

  \midrule

  Qwen2.5-7B & MATH500
  & 26.40{\scriptsize $\pm$0.35} & 20.3{\scriptsize $\pm$0.4}
  & 37.53{\scriptsize $\pm$1.37} & 33.3{\scriptsize $\pm$1.8}
  & 34.20{\scriptsize $\pm$0.72} & \textbf{38.7{\scriptsize $\pm$1.7}}
  & \cellcolor{lightgreen}\textbf{38.67{\scriptsize $\pm$0.90}} & \cellcolor{lightgreen}33.9{\scriptsize $\pm$1.3}
  & 1.0$\times$ & 1.6$\times$ & \textbf{1.9$\times$} & \cellcolor{lightgreen}1.7$\times$ \\

  & GSM8K
  & 53.53{\scriptsize $\pm$0.26} & 20.1{\scriptsize $\pm$0.2}
  & \textbf{72.25{\scriptsize $\pm$3.43}} & 29.5{\scriptsize $\pm$0.8}
  & 69.02{\scriptsize $\pm$0.93} & \textbf{42.4{\scriptsize $\pm$1.1}}
  & \cellcolor{lightgreen}71.24{\scriptsize $\pm$1.18} & \cellcolor{lightgreen}38.2{\scriptsize $\pm$0.2}
  & 1.0$\times$ & 1.5$\times$ & \textbf{2.1$\times$} & \cellcolor{lightgreen}1.9$\times$ \\

  \midrule

  Llama-3-8B & MATH500
  & 13.33{\scriptsize $\pm$0.75} & 17.8{\scriptsize $\pm$0.2}
  & \textbf{20.07{\scriptsize $\pm$0.09}} & 27.1{\scriptsize $\pm$0.9}
  & 17.93{\scriptsize $\pm$0.12} & \textbf{33.2{\scriptsize $\pm$2.1}}
  & \cellcolor{lightgreen}19.27{\scriptsize $\pm$0.41} & \cellcolor{lightgreen}27.8{\scriptsize $\pm$1.0}
  & 1.0$\times$ & 1.5$\times$ & \textbf{1.9$\times$} & \cellcolor{lightgreen}1.6$\times$ \\

  & GSM8K
  & 44.00{\scriptsize $\pm$2.18} & 18.4{\scriptsize $\pm$0.2}
  & \textbf{56.46{\scriptsize $\pm$0.58}} & 25.6{\scriptsize $\pm$0.3}
  & 54.31{\scriptsize $\pm$0.04} & \textbf{33.9{\scriptsize $\pm$1.7}}
  & \cellcolor{lightgreen}56.10{\scriptsize $\pm$0.86} & \cellcolor{lightgreen}25.3{\scriptsize $\pm$0.4}
  & 1.0$\times$ & 1.4$\times$ & \textbf{1.8$\times$} & \cellcolor{lightgreen}1.4$\times$ \\

  \bottomrule
  \end{tabular}}
\end{table*}

\begin{figure}[t]
	\centering
	\begin{subfigure}{0.32\textwidth}
		\centering
		\includegraphics[width=\linewidth]{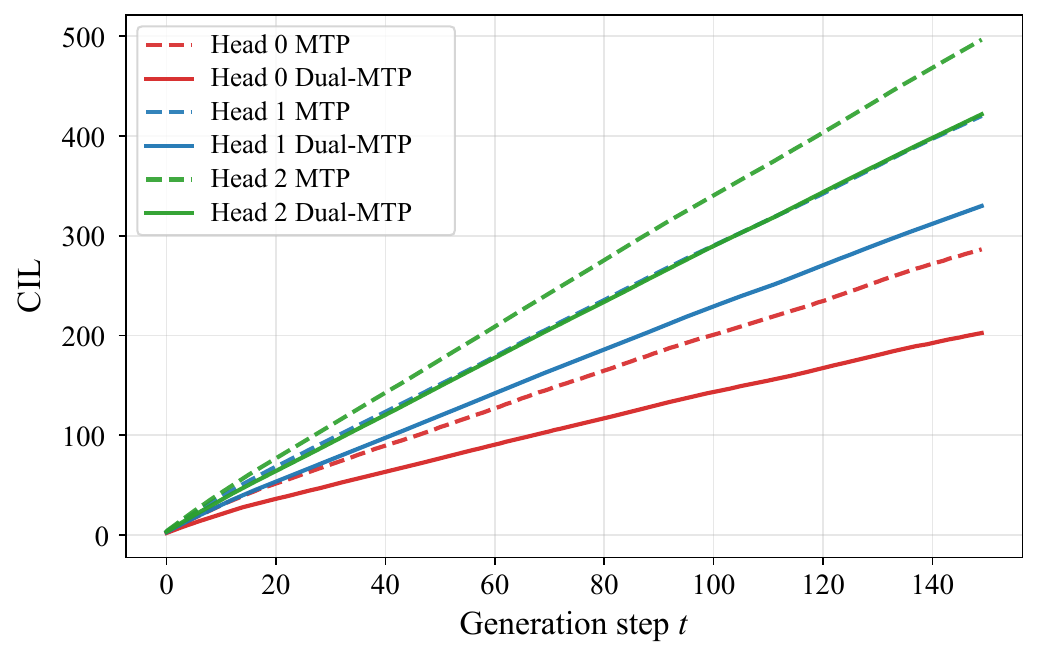}
		\caption{Qwen2.5-7B.}
	\end{subfigure}
	\hfill
	\begin{subfigure}{0.32\textwidth}
		\centering
		\includegraphics[width=\linewidth]{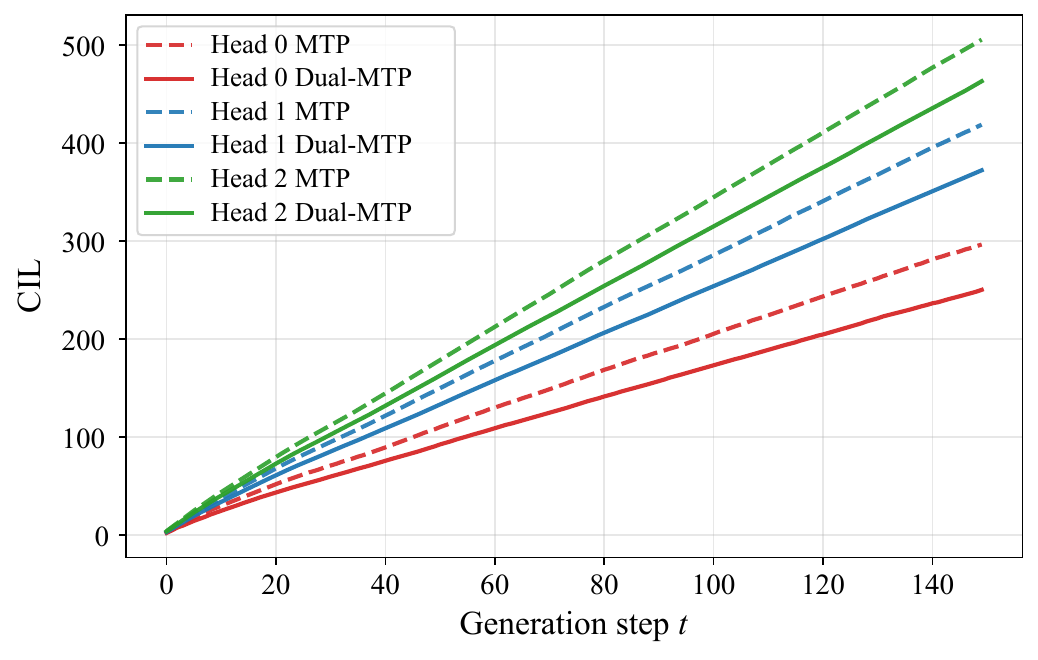}
		\caption{Llama-3-8B.}
	\end{subfigure}
	\hfill
	\begin{subfigure}{0.32\textwidth}
		\centering
		\includegraphics[width=\linewidth]{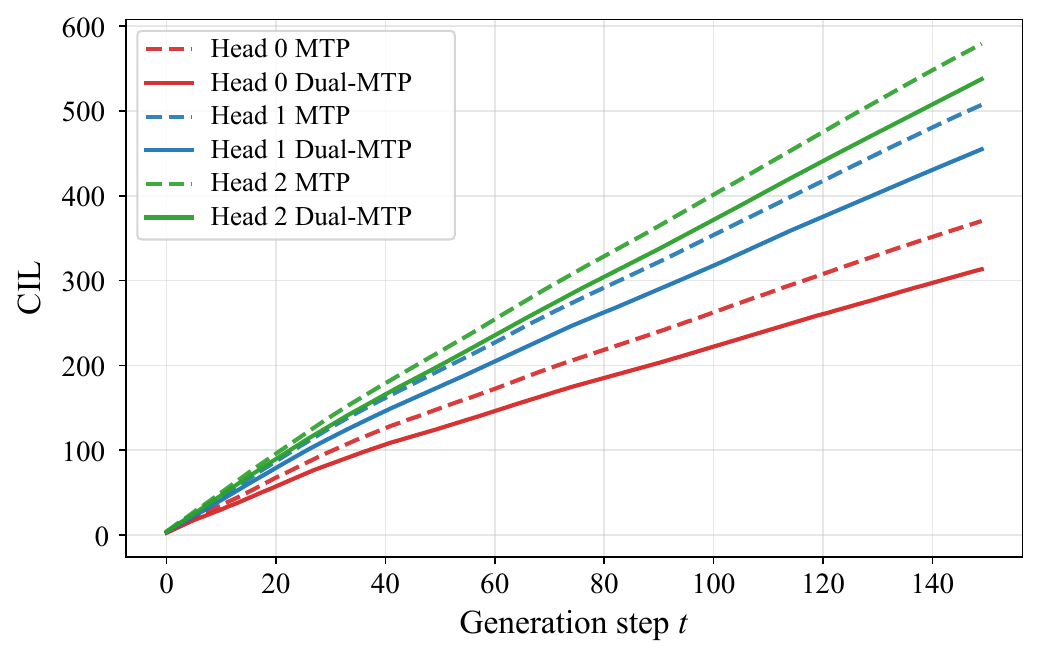}
		\caption{DeepSeek-R1-Distill-Qwen-7B.}
	\end{subfigure}
    \caption{
\textbf{Cumulative predictive entropy across draft heads.} Dual-MTP exhibits lower predictive entropy than MTP across the evaluated heads and backbones.
}
	\label{fig:cil_heads}
	\vspace{-10pt}
\end{figure}

\begin{figure}[t]
	\centering
	\includegraphics[width=\linewidth]{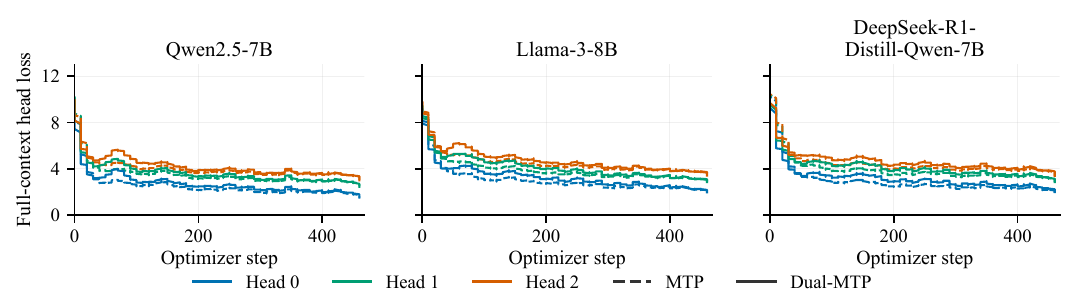}
	\caption{Per-head full-context training loss on GSM8K. Curves use a 15-observation moving average; panels share both axes. Dashed lines denote MTP and solid lines denote Dual-MTP.}
	\label{fig:loss_curves}
	\vspace{-10pt}
\end{figure}

\begingroup{}
\textbf{Predictive Entropy across Heads.} \Cref{fig:cil_heads} compares the cumulative predictive-entropy proxy $\widehat{\mathrm{CIL}}_{<l,k}$, obtained by summing each head's predictive entropy along the generated trajectory (\cref{app:empirical-cil}). The reference-target conditional distribution is unavailable along a generated trajectory, so the plotted statistic uses the model distribution. Its relation to reference-target uncertainty depends on calibration; we therefore interpret it together with per-head prediction accuracy. Predictive entropy increases with head index. Dual-MTP lowers it at each evaluated head and backbone, with Head~0 reductions of $29.3\%$ on Qwen2.5-7B, $15.6\%$ on Llama-3-8B, and $15.3\%$ on \mbox{DeepSeek-R1-Distill-Qwen-7B}.
\endgroup{}

\begin{wrapfigure}[14]{r}{0.45\textwidth}
\vspace{-\intextsep}
\centering
\includegraphics[width=\linewidth]{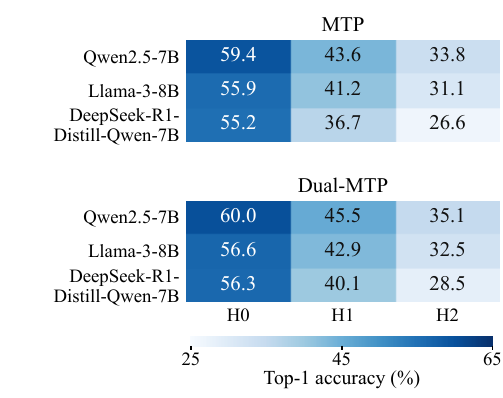}
\captionsetup{justification=raggedright,singlelinecheck=false}
\caption{GSM8K top-1 accuracy by head.}
\label{fig:head_accuracy}
\end{wrapfigure}

\leavevmode
\textbf{Per-Head Accuracy and Training Dynamics.} \Cref{fig:head_accuracy,fig:loss_curves} complement the predictive-entropy view of \cref{fig:cil_heads} on GSM8K. 
\Cref{fig:loss_curves} shows per-head training-loss curves. Dual-MTP's mean full-context training loss is slightly higher than MTP's at each head. The mean loss gap between Head~0 and Head~2 is similar for the two methods on Qwen2.5-7B and slightly smaller under Dual-MTP on Llama-3-8B and DeepSeek-R1-Distill-Qwen-7B.
\Cref{fig:head_accuracy} reports per-head Avg@1 accuracy. MTP degrades sharply with head index, mirroring the monotonic CIL growth, while Dual-MTP retains higher accuracy on every head, with the largest gain at Head~1 on all three backbones. 
For example, on Qwen2.5-7B, Dual-MTP improves token prediction accuracy by 0.63, 1.92, and 1.26 percentage points at Heads~0, 1, and~2, respectively. For matched head--backbone pairs, Dual-MTP combines lower predictive entropy with higher head-level accuracy.

\Needspace{18\baselineskip}
\leavevmode\textbf{Code Generation.} We evaluate Qwen2.5-7B-Instruct on HumanEval+ and MBPP+ using one solution per problem. Each method trains its draft heads on greedy responses from its frozen verifier. On HumanEval+, Dual-MTP achieves $76.63\%$ pass@1 at $K_v=1$, compared with $73.98\%$ for Medusa and $73.37\%$ for MTP (\cref{tab:code-generation-results}). On MBPP+, MTP is the closest competitor. Its gap to Dual-MTP is only $0.35$ points at $K_v=1$ but exceeds $2.6$ points at both $K_v=3$ and $5$. Thus, Dual-MTP has the highest mean pass@1 across all six comparisons, with its clearest advantage on MBPP+ under relaxed verification. Appendix~\ref{app:code-generation} gives the evaluation protocol.

\leavevmode\textbf{Efficiency Analysis.}
\Cref{tab:decoding_speedup} shows wall-clock throughput on MATH500 and GSM8K relative to NTP. Dual-MTP accelerates NTP decoding by $1.4\times$–$1.9\times$ across the five backbones and surpasses MTP on $8$ of $10$ settings, with the largest gain on Qwen2.5-7B/GSM8K ($38.20$ tok/s, $1.90\times$ vs.\ MTP's $1.47\times$). 
Across all ten math settings, Medusa decodes faster than Dual-MTP, while Dual-MTP achieves higher accuracy.
Appendix~\ref{app:accept_rate} reports per-task speculative acceptance rates across all five evaluated backbones.

\begin{figure}[t]
\centering
\includegraphics[width=\linewidth]{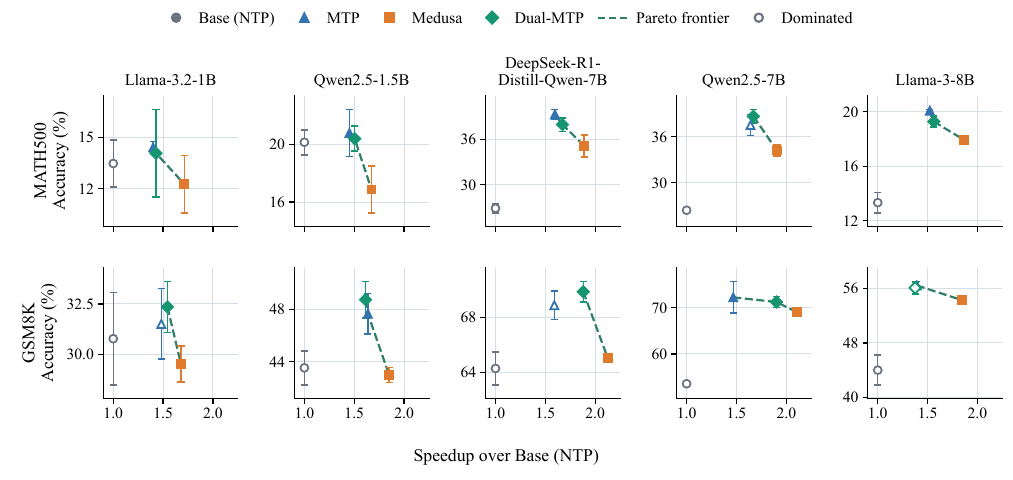}
\caption{Math accuracy and speedup over NTP at $K_v=5$ (\cref{tab:decoding_speedup}). Bars show accuracy SD. Dashed lines connect nondominated means. Hollow markers denote methods matched or exceeded on both axes and surpassed on at least one.}
\label{fig:math_pareto}
\end{figure}

\leavevmode\textbf{Accuracy--Speed Trade-Off.} Dual-MTP lies on the accuracy--speed frontier in nine of the ten settings in \cref{fig:math_pareto}. On Qwen2.5-7B/GSM8K, MTP achieves the highest accuracy ($72.25\%$), and Medusa decodes fastest ($2.11\times$ NTP). Dual-MTP falls between them at $71.24\%$ and $1.90\times$ NTP. On MATH500, MTP is more accurate than Dual-MTP on four backbones, but Dual-MTP is faster on all four. MTP leads on both metrics only on GSM8K with Llama-3-8B. On HumanEval+, relaxing verification from $K_v=1$ to 3 raises Dual-MTP\textquotesingle s speedup from $1.364\times$ to $1.532\times$ but lowers pass@1 from $76.63\%$ to $74.19\%$ (\cref{tab:code-generation-results}). On MBPP+, speedup rises from $1.344\times$ to $1.481\times$ as pass@1 falls from $70.46\%$ to $69.22\%$. At $K_v=5$, speed changes little on either task.

\begin{table}[!t]
\centering\small
\caption{Code generation on Qwen2.5-7B-Instruct. Cells report EvalPlus pass@1 (\%, mean $\pm$ sample SD) and mean speedup over fine-tuned NTP across three independent training runs.}
\label{tab:code-generation-results}
\setlength{\tabcolsep}{3pt}
\begin{tabular}{llccc}
\toprule
Task & Method & $K_v=1$ & $K_v=3$ & $K_v=5$\\
\midrule
HumanEval+ & MTP & $73.37\pm0.93$ / 1.304 & $71.75\pm1.53$ / 1.489 & $71.34\pm2.20$ / 1.483\\
& Medusa & $73.98\pm1.27$ / 1.301 & $72.97\pm3.01$ / 1.469 & $72.97\pm3.57$ / 1.467\\
& Dual-MTP & $76.63\pm0.35$ / 1.364 & $74.19\pm0.93$ / 1.532 & $74.19\pm0.93$ / 1.530\\
\midrule
MBPP+ & MTP & $70.11\pm0.00$ / 1.365 & $66.58\pm1.07$ / 1.503 & $66.23\pm0.40$ / 1.518\\
& Medusa & $69.40\pm0.67$ / 1.366 & $65.61\pm1.15$ / 1.503 & $65.34\pm1.40$ / 1.498\\
& Dual-MTP & $70.46\pm0.40$ / 1.344 & $69.22\pm1.10$ / 1.481 & $69.05\pm1.73$ / 1.486\\
\bottomrule
\end{tabular}
\end{table}

\Needspace{5\baselineskip}
\subsection{Ablation Studies}
\label{subsec:ablation}

\begingroup
\setlength{\emergencystretch}{2em}
\tolerance=2000
\leavevmode
\textbf{Contribution of Full-Context Supervision.} We evaluate Mask-only training on Qwen2.5-7B/1.5B-Instruct over GSM8K and MATH500, disabling both NTP and full-context head losses while retaining the corresponding Dual-MTP masking settings. Mask-only accuracy is $0.00\%$--$0.23\%$, compared with $20.40\%$--$71.24\%$ for Dual-MTP. On Qwen2.5-7B, GSM8K accuracy is $0.10\%$ versus $71.24\%$, and MATH500 accuracy is $0.07\%$ versus $38.67\%$. This degradation supports retaining full-context supervision when optimizing the backbone and draft heads with masked contexts. Appendix~\ref{app:ablation-mask-only} gives the configurations and reference comparisons.

\textbf{Effect of Draft-Head Count.} We evaluate $D\in\{1,2,3,4,5\}$ on Qwen2.5-7B-Instruct and Qwen2.5-1.5B-Instruct using all 1,319 GSM8K questions. Across ten backbone--head-count pairs, Dual-MTP exceeds MTP in accuracy in three cases, with differences from $-1.82$ to $+1.09$ percentage points. Acceptance rate is higher in nine cases, with a mean increase of $3.42$ points and a range of $-0.08$ to $+15.77$ points. Higher draft acceptance therefore does not consistently imply higher task accuracy. Appendix~\ref{app:ablation-h-sweep} reports the full results and the training configurations.
\par
\endgroup

\section{Conclusion}

We introduced \textbf{Dual-MTP}, which trains shared prediction heads under full and randomly masked contexts while leaving the MTP architecture and inference procedure unchanged.
Our analysis characterizes the reference-target block-sequence CIL gap and uses the normalized dual loss to bound the expected Jensen--Shannon divergence between predictions under full and masked contexts.
Across five backbones and six benchmarks spanning math reasoning and GLUE understanding, Dual-MTP attains the highest average task score on four backbones, with gains of up to $2.46$ points over MTP. It outperforms Medusa on average across the six tasks on every backbone, by up to $4.87$ points. Original throughput is $1.4\times$--$1.9\times$ NTP.
Head-level accuracy and predictive-entropy measurements provide complementary evidence of improved prediction quality, while the verification-parameter experiments illustrate the \mbox{accuracy--speed trade-off}.

\section{\texorpdfstring{Limitations}{Limitations}}
\label{sec:limitations}

Dual-MTP has lower inference throughput than Medusa in the ten mathematical-reasoning settings of \cref{tab:decoding_speedup}. Further work on draft generation and verification could reduce decoding latency and improve overall throughput. Our evaluation covers backbones ranging from 1B to 9B parameters. Extending the experiments to larger backbones would allow us to examine inference speed and memory requirements at larger model scales.

\clearpage
\renewcommand{\acksection}{\section*{Acknowledgments}}
\begin{ack}
TYJ, ZKZ, and BH were supported by RGC General Research Fund No.\ 12202026, RGC Research Fellow Scheme No.\ RFS2627-2S02, RGC Young Collaborative Research Grant No.\ C2005-24Y, RGC General Research Fund No.\ 12200725, NSFC Major Research Plan No.\ 92570109, NSFC General Program No.\ 62376235, Guangdong and Hong Kong Universities ``1+1+1'' Joint Research Collaboration Scheme, HKBU Faculty Niche Research Areas No.\ RC-FNRA-IG/25-26/SCI/05, and HKBU CSD Departmental Incentive Scheme. This work is supported in part by Guangdong and Hong Kong Universities ``1+1+1'' Joint Research Collaboration Scheme under Grant JRCS-24-25-COMP-03, and the RGC Senior Research Fellow Scheme under Grant SRFS2324-2S02. XHJ and HDW were supported by HK RGC research grant No.: SRFS2425-1S01.
\end{ack}
{
    \small
\let\originalbibitem\bibitem
\renewcommand{\bibitem}[2][]{%
  \par\normalcolor
  \originalbibitem[#1]{#2}%
}
\bibliographystyle{plainnat}
\bibliography{references}

}

\appendix
\crefalias{section}{appendix}
\crefalias{subsection}{appendix}
\crefalias{subsubsection}{appendix}
\begingroup
\clubpenalty=10000
\widowpenalty=10000
\displaywidowpenalty=10000
\begingroup{}
\section{Proof of Theorem~\ref{thm:main}}\label{app:1}
\label{app:proof-thm1}

\subsection{Reference Targets and Generated Predictions}\label{app:setup}
The training procedure produces parameters $\theta$, which are held fixed throughout the inference analysis. A test input and reference target are drawn from $p_{\mathrm{data}}(X,T)$. We fix a finite horizon and a block partition; end-of-sequence padding handles shorter outputs. Write $T=(T_1^B,\ldots,T_M^B)$ and $R_N=(R_1^B,\ldots,R_M^B)$. Each block in the analysis consists of a fixed group of output positions.

\subsection{Decoding Assumption}\label{app:assumption}
Let $\pi_\theta$ denote the conditional distribution over the next output block induced by the decoding procedure. The decoder has access only to $X$, its preceding predictions, and randomness independent of the reference target given $X$. Thus
\begin{equation}\label{eq:assumption-A1}
 p(R_N\mid X,T)=\prod_{m=1}^M\pi_\theta(R_m^B\mid X,R_{<m}^B)=p(R_N\mid X),
\end{equation}
and hence
\begin{equation}\label{eq:RN-indep}
 R_N\perp T\mid X.
\end{equation}
Predictions and targets share the input $X$. Conditional on $X$, the decoder generates $R_N$ using its own predictions and independent randomness, which gives \cref{eq:RN-indep}.

\subsection{Proof}
\begin{proof}
Conditional independence implies
\begin{equation}\label{eq:Rm-indep}
H(T_m^B\mid R_m^B,X)=H(T_m^B\mid X),\qquad
H(T\mid R_N,X)=H(T\mid X).
\end{equation}
Substitution into the definition of $\Delta$ and the conditional entropy chain rule give
\begin{align}\label{eq:delta-decomposition}
\Delta
&=\sum_{m=1}^{M}H(T_m^B\mid X)-H(T\mid X)\notag\\
&=\sum_{m=1}^{M}\bigl[H(T_m^B\mid X)-H(T_m^B\mid T_{<m}^B,X)\bigr]\notag\\
&=\sum_{m=1}^{M}I(T_m^B;T_{<m}^B\mid X).
\end{align}
\end{proof}

\paragraph{Remark.}
For a fixed reference distribution and partition, the input-conditional gap is invariant to the trained parameters. It equals zero when the reference target is deterministic given $X$. Thus, the identity characterizes the dependence structure of the reference targets; the loss-dependent analysis below characterizes predictive consistency across context views.

\section{Proof of Theorem~\ref{thm:view-mi}}\label{app:view-mi}

\paragraph{Definitions and Normalization.}
For the same training example, mask, head, and supervised position, let $Y$ be the target token and $q_f,q_m$ the predictive distributions under full and masked contexts. With $q_a=(q_f+q_m)/2$ and natural logarithms, their Jensen--Shannon divergence is
\begin{equation}
\operatorname{JS}(q_f,q_m)
=\tfrac12\operatorname{KL}(q_f\Vert q_a)+\tfrac12\operatorname{KL}(q_m\Vert q_a).
\end{equation}
We use the same normalized weights over examples, masks, heads, and positions in all expectations. The sum of the two mean token losses is
\begin{equation}
J=\mathbb E[-\ln q_f(Y)-\ln q_m(Y)]
=\overline{\mathcal L}_{\mathrm{full}}+\overline{\mathcal L}_{\mathrm{masked}}.
\end{equation}

\paragraph{A Pointwise Bound.}
For two distributions $p,q$ on the token vocabulary, set $a_z=(p_z+q_z)/2$ and $d_z=|p_z-q_z|/(p_z+q_z)$ when $a_z>0$. Their contribution to Jensen--Shannon divergence is $a_z f(d_z)$, where
\[
f(d)=\tfrac12\bigl[(1+d)\ln(1+d)+(1-d)\ln(1-d)\bigr].
\]
Since $f$ is convex on $[0,1]$, $f(0)=0$, and $f(1)=\ln2$, it lies below the chord $d\ln2$. Consequently,
\[
\operatorname{JS}(p,q)\le(\ln2)\sum_z a_zd_z
=(\ln2)\operatorname{TV}(p,q).
\]
Also $\operatorname{TV}(p,q)=1-\sum_z\min\{p_z,q_z\}\le1-\min\{p_y,q_y\}$ for any target token $y$. Since $p_yq_y\le\min\{p_y,q_y\}$, setting $\ell=-\ln q_f(y)-\ln q_m(y)$ gives
\begin{align}
\operatorname{JS}(q_f,q_m)
&\le(\ln2)\operatorname{TV}(q_f,q_m)\notag\\
&\le(\ln2)\bigl(1-\min\{q_f(y),q_m(y)\}\bigr)\notag\\
&\le(\ln2)\bigl(1-q_f(y)q_m(y)\bigr)
=(\ln2)(1-e^{-\ell}).
\end{align}
Zero target probability is handled by the limit $\ell=+\infty$.

\paragraph{Averaging and the Weighted Objective.}
The function $u\mapsto1-e^{-u}$ is concave. Averaging the pointwise inequality over the same examples, masks, heads, and token positions and applying Jensen's inequality proves
\[
\mathbb E[\operatorname{JS}(q_f,q_m)]\le(\ln2)\mathbb E[1-e^{-\ell}]
\le(\ln2)(1-e^{-\mathbb E\ell})=(\ln2)(1-e^{-J}).
\]
The upper envelope has derivative $(\ln2)e^{-J}>0$. Hence, reducing $J$ tightens the bound, and $J\to0$ implies $\mathbb E[\operatorname{JS}(q_f,q_m)]\to0$.

Fix any training step $s$ with $\lambda(s)>0$. Nonnegative cross-entropies give
\[
L_{\lambda(s)}
=\overline{\mathcal L}_{\mathrm{full}}+\lambda(s)\,\overline{\mathcal L}_{\mathrm{masked}}
\ge\min\{1,\lambda(s)\}J.
\]
Consequently, $J\le L_{\lambda(s)}/\min\{1,\lambda(s)\}$. Monotonicity of the preceding upper envelope yields
\[
\mathbb E[\operatorname{JS}(q_f,q_m)]
\le(\ln2)(1-e^{-J})
\le(\ln2)\left[1-\exp\!\left(-\frac{L_{\lambda(s)}}{\min\{1,\lambda(s)\}}\right)\right],
\]
which proves \cref{thm:view-mi}. The inequality holds pointwise in $s$, allowing $\lambda(s)$ to vary over the warm-up schedule. For $\lambda(s)=1$, $L_{\lambda(s)}=J$ and the unweighted bound is recovered.

Prediction consistency follows when $L_{\lambda(s)}/\min\{1,\lambda(s)\}\to0$. In particular, if $\lambda(s)$ is uniformly bounded below by a positive constant, $L_{\lambda(s)}\to0$ suffices. At a step with $\lambda(s)=0$, the weighted objective contains only full-context supervision and the weighted bound does not apply; the unweighted bound in terms of $J$ remains valid.

\endgroup{}

\section{Token Stochastic Masking: Implementation Details}
\label{app:token-masking}

This appendix provides the full specification of the token stochastic masking mechanism described in \cref{subsec:dual-training}. At each training step, a single binary mask over token positions is sampled and shared across all prediction heads. This design constrains access to historical context during training while preserving standard autoregressive inference behavior, since masking is disabled at inference time.

\subsection{Annealed Masking Rate}

We adopt an annealing schedule to dynamically adjust the masking rate throughout training. Let $s$ denote the current training step and $S$ the maximum number of steps. We define the training progress as $p = \min(s / S, 1)$. The base masking rate is computed via exponential decay:
\begin{equation}
\tilde{\gamma}(p) = \gamma_{\text{final}} + (\gamma_{\text{init}} - \gamma_{\text{final}})\exp(-\alpha p),
\end{equation}
where $\gamma_{\text{init}}$ and $\gamma_{\text{final}}$ are the initial and final mask rates and $\alpha$ controls the decay.

To encourage exploration early in training, we further introduce temperature-controlled stochastic perturbations:
\begin{equation}
\gamma = \tilde{\gamma}(p) + \epsilon, \qquad \epsilon \sim \mathcal{N}\!\left(0,\sigma^2(p)\right),
\end{equation}
where the variance $\sigma^2(p)$ is proportional to a temperature parameter that decays over training. The resulting masking rate is clipped to a predefined range to ensure stability.

\subsection{Mask Construction and Application}

Given a masking rate $\gamma$, we sample a binary mask over token positions as
\begin{equation}
M_i \sim \mathrm{Bernoulli}(1 - \gamma).
\end{equation}
The sampled mask is incorporated into the attention computation. Concretely, for the attention logits $A^{(h)} \in \mathbb{R}^{Q \times K}$ corresponding to head $h$, we apply the transformation
\begin{equation}
A^{(h)}_{q,i} \leftarrow
\begin{cases}
A^{(h)}_{q,i}, & M_i = 1, \\
-\infty, & M_i = 0,
\end{cases}
\end{equation}
prior to the softmax operation. This prevents queries in the current block from attending to masked tokens. All attention heads use the same mask at each training step.

\begingroup{}
\section{Empirical Predictive-Entropy Proxy}
\label{app:empirical-cil}

\subsection{Definition}
For head $k$ at generation position $i$, let $h_i$ be the recorded hidden state and $q_{i,k}=P_{\theta_k}(\cdot\mid h_i)$ its vocabulary distribution. The plotted quantity is
\begin{gather}
\widehat H_{i,k}=H(q_{i,k})=-\sum_{x\in\mathcal V}q_{i,k}(x)\ln q_{i,k}(x),\label{eq:emp_cil}\\
\widehat{\mathrm{CIL}}_{<l,k}=\sum_{i=1}^{l-1}\widehat H_{i,k}.\label{eq:emp_cum_cil}
\end{gather}
This statistic measures cumulative uncertainty in the model distributions along a generated trajectory. It conditions on hidden states $h_i$, while the reference-target CIL in \cref{eq:cum_ntp} conditions on predicted tokens $r_i$.

\subsection{Calibration-Dependent Interpretation}\label{app:cil-relation}
Let $p_i=p_{\mathrm{data}}(T_i\in\cdot\mid h_i)$ denote the conditional reference distribution at the same inference states, and define the state-conditioned quantity $\mathrm{CIL}^{\mathrm{state}}_{<l}=\sum_{i<l}H(p_i)$. The following result relates the predictive-entropy statistic to this reference quantity through the discrepancy between $p_i$ and $q_{i,k}$.

\begin{proposition}[State-conditioned surrogate]\label{prop:cil-relation}
For a finite vocabulary of size $V\ge2$, let $\delta_{i,k}=\tfrac12\|p_i-q_{i,k}\|_1$. If $\delta_{i,k}\le1-1/V$ at each position, then
\begin{equation}\label{eq:cil-gap}
\left|\widehat{\mathrm{CIL}}_{<l,k}-\mathrm{CIL}^{\mathrm{state}}_{<l}\right|
\le\sum_{i<l}\left[\delta_{i,k}\ln(V-1)+h_2(\delta_{i,k})\right],
\end{equation}
where $h_2$ is binary entropy with natural logarithms. Equality of the two cumulative quantities follows if $p_i=q_{i,k}$ for every position. For a fixed horizon, the difference vanishes if the forward KL errors tend to zero at all these states.
\end{proposition}
\begin{proof}
Apply the finite-alphabet entropy continuity bound to each pair $(p_i,q_{i,k})$ and then the triangle inequality~\cite{cover2006elements}. Pinsker's inequality gives $\delta_{i,k}\le\sqrt{\operatorname{KL}(p_i\Vert q_{i,k})/2}$, so convergence of the state-wise KL errors and continuity of the right-hand side prove the final statement.
\end{proof}
The approximation quality depends on distributional agreement along the model\textquotesingle s generated trajectories. We therefore evaluate predictive entropy jointly with task and head-level accuracy to distinguish more accurate predictions from overconfidence.
\endgroup{}

\section{Speculative Acceptance Rate Analysis}
\label{app:accept_rate}

\Cref{tab:accept_rate} compares the fraction of draft tokens accepted by the verifier. Dual-MTP generally achieves higher or comparable acceptance rates to MTP, indicating closer agreement between its draft predictions and verifier decisions.
Medusa achieves higher average acceptance rates on most backbones, while Dual-MTP achieves higher average task scores on all five backbones (\cref{tab:accept_rate,tab:main_results}).
The methods use the same verification rule at a given threshold; differences in acceptance arise from their learned draft and verifier distributions.
Together, \cref{tab:main_results,tab:decoding_speedup,tab:accept_rate} characterize prediction accuracy, accepted draft length, and decoding throughput.

\begin{table*}[t]
\centering
\caption{Speculative decoding acceptance rate (mean $\pm$ SD over three decoding runs).}
\resizebox{\linewidth}{!}{
\begin{tabular}{c|l|cc|cccc|c}
\toprule
Model & Setting & Math500 & GSM8K & CoLA & MRPC & RTE & WNLI & \textbf{Avg. $\uparrow$} \\
\midrule

\multirow{3}{*}{\makecell[c]{Llama-3.2-1B\\-Instruct}}
& MTP & 49.1{\scriptsize $\pm$0.3} & 52.5{\scriptsize $\pm$0.5} & 57.1{\scriptsize $\pm$1.7} & 64.1{\scriptsize $\pm$6.1} & 50.0{\scriptsize $\pm$0.0} & 50.0{\scriptsize $\pm$0.0} & 53.8 \\
& Medusa & \textbf{56.5{\scriptsize $\pm$0.7}} & \textbf{62.0{\scriptsize $\pm$0.6}} & \textbf{59.2{\scriptsize $\pm$0.2}} & \textbf{65.6{\scriptsize $\pm$0.7}} & \textbf{50.0{\scriptsize $\pm$0.0}} & 50.0{\scriptsize $\pm$0.0} & \textbf{57.2} \\
& \cellcolor{lightgreen}Dual-MTP & \cellcolor{lightgreen}49.1{\scriptsize $\pm$0.1} & \cellcolor{lightgreen}52.6{\scriptsize $\pm$0.6} & \cellcolor{lightgreen}56.3{\scriptsize $\pm$1.8} & \cellcolor{lightgreen}64.6{\scriptsize $\pm$6.3} & \cellcolor{lightgreen}50.0{\scriptsize $\pm$0.0} & \cellcolor{lightgreen}50.0{\scriptsize $\pm$0.0} & \cellcolor{lightgreen}53.8 \\

\midrule

\multirow{3}{*}{\makecell[c]{Qwen2.5-1.5B\\-Instruct}}
& MTP & 47.7{\scriptsize $\pm$0.3} & 49.8{\scriptsize $\pm$0.4} & \textbf{66.3{\scriptsize $\pm$0.0}} & 61.2{\scriptsize $\pm$0.3} & 44.9{\scriptsize $\pm$0.1} & \textbf{66.2{\scriptsize $\pm$0.5}} & 56.0 \\
& Medusa & \textbf{55.0{\scriptsize $\pm$0.4}} & \textbf{59.5{\scriptsize $\pm$0.4}} & 54.0{\scriptsize $\pm$0.2} & \textbf{62.2{\scriptsize $\pm$0.7}} & \textbf{45.1{\scriptsize $\pm$0.4}} & 65.5{\scriptsize $\pm$0.3} & \textbf{56.9} \\
& \cellcolor{lightgreen}Dual-MTP & \cellcolor{lightgreen}48.3{\scriptsize $\pm$0.4} & \cellcolor{lightgreen}50.4{\scriptsize $\pm$0.4} & \cellcolor{lightgreen}56.2{\scriptsize $\pm$0.7} & \cellcolor{lightgreen}61.1{\scriptsize $\pm$0.4} & \cellcolor{lightgreen}44.9{\scriptsize $\pm$0.1} & \cellcolor{lightgreen}63.2{\scriptsize $\pm$0.1} & \cellcolor{lightgreen}54.0 \\

\midrule

\multirow{3}{*}{\makecell[c]{DeepSeek\\-R1-Distill-Qwen-7B}}
& MTP & 52.0{\scriptsize $\pm$0.7} & 55.2{\scriptsize $\pm$0.3} & \textbf{26.5{\scriptsize $\pm$0.3}} & 43.2{\scriptsize $\pm$1.0} & \textbf{12.7{\scriptsize $\pm$0.2}} & \textbf{34.1{\scriptsize $\pm$0.8}} & 37.3 \\
& Medusa & \textbf{60.0{\scriptsize $\pm$0.6}} & \textbf{65.6{\scriptsize $\pm$0.2}} & 24.1{\scriptsize $\pm$0.2} & 29.4{\scriptsize $\pm$1.4} & 11.9{\scriptsize $\pm$0.4} & 32.9{\scriptsize $\pm$0.5} & 37.3 \\
& \cellcolor{lightgreen}Dual-MTP & \cellcolor{lightgreen}52.3{\scriptsize $\pm$0.5} & \cellcolor{lightgreen}58.5{\scriptsize $\pm$0.4} & \cellcolor{lightgreen}24.2{\scriptsize $\pm$0.6} & \cellcolor{lightgreen}\textbf{44.8{\scriptsize $\pm$1.1}} & \cellcolor{lightgreen}11.8{\scriptsize $\pm$1.8} & \cellcolor{lightgreen}32.0{\scriptsize $\pm$0.5} & \cellcolor{lightgreen}\textbf{37.3} \\
\midrule

\multirow{3}{*}{\makecell[c]{Qwen2.5-7B\\-Instruct}}
& MTP & 53.2{\scriptsize $\pm$0.3} & 41.2{\scriptsize $\pm$0.7} & 65.7{\scriptsize $\pm$0.5} & \textbf{87.6{\scriptsize $\pm$0.8}} & \textbf{61.9{\scriptsize $\pm$2.0}} & 57.1{\scriptsize $\pm$0.5} & 61.1 \\
& Medusa & \textbf{61.2{\scriptsize $\pm$0.2}} & \textbf{65.5{\scriptsize $\pm$0.1}} & \textbf{71.1{\scriptsize $\pm$0.1}} & 69.4{\scriptsize $\pm$0.2} & 49.3{\scriptsize $\pm$0.0} & 71.4{\scriptsize $\pm$0.2} & 64.7 \\
& \cellcolor{lightgreen}Dual-MTP & \cellcolor{lightgreen}53.2{\scriptsize $\pm$0.2} & \cellcolor{lightgreen}57.1{\scriptsize $\pm$0.1} & \cellcolor{lightgreen}66.8{\scriptsize $\pm$0.3} & \cellcolor{lightgreen}85.9{\scriptsize $\pm$0.4} & \cellcolor{lightgreen}61.0{\scriptsize $\pm$0.6} & \cellcolor{lightgreen}\textbf{72.7{\scriptsize $\pm$0.9}} & \cellcolor{lightgreen}\textbf{66.1} \\
\midrule

\multirow{3}{*}{\makecell[c]{Llama-3-8B\\-Instruct}}
& MTP & 53.4{\scriptsize $\pm$0.1} & 40.7{\scriptsize $\pm$0.4} & 54.2{\scriptsize $\pm$0.7} & \textbf{73.3{\scriptsize $\pm$1.9}} & 50.0{\scriptsize $\pm$0.0} & 50.8{\scriptsize $\pm$0.4} & 53.7 \\
& Medusa & \textbf{60.4{\scriptsize $\pm$0.3}} & \textbf{67.9{\scriptsize $\pm$0.6}} & \textbf{55.0{\scriptsize $\pm$0.0}} & 72.4{\scriptsize $\pm$0.1} & \textbf{50.0{\scriptsize $\pm$0.0}} & 50.0{\scriptsize $\pm$0.0} & \textbf{59.3} \\
& \cellcolor{lightgreen}Dual-MTP & \cellcolor{lightgreen}55.3{\scriptsize $\pm$0.6} & \cellcolor{lightgreen}62.9{\scriptsize $\pm$0.4} & \cellcolor{lightgreen}54.4{\scriptsize $\pm$0.6} & \cellcolor{lightgreen}71.6{\scriptsize $\pm$3.9} & \cellcolor{lightgreen}50.0{\scriptsize $\pm$0.0} & \cellcolor{lightgreen}\textbf{50.8{\scriptsize $\pm$0.2}} & \cellcolor{lightgreen}57.5 \\

\bottomrule
\end{tabular}
}
\label{tab:accept_rate}
\vspace{-8pt}
\end{table*}

\section{Ablation Studies: Full Results}
\label{app:ablation}

We compare training objectives, the number of draft heads, and the warm-up schedule for the masked loss. All three ablations use $K_v=5$.

\paragraph{Scoring.}\label{app:final-answer}
All mathematical-reasoning accuracies use the same final-answer rule on at most 2048 response tokens, excluding the prompt. We extract the final answer from a boxed expression, an answer label, a concluding statement, a final equation, or a standalone answer. We count it as correct if it is mathematically equivalent to the reference answer. Missing, incomplete, unparseable, or contradictory final answers count as incorrect; occurrences of the reference within intermediate reasoning do not establish correctness. Historical responses are truncated before rescoring, while new evaluations enforce the limit during generation. In Tables~\ref{tab:main_results}, \ref{tab:decoding_speedup}, and \ref{tab:ablation-mask-only}, SD is the sample SD for 1B/1.5B models and population SD for 7B/8B models, over repeated decoding runs. Other repeated mathematical evaluations report sample SD; SD is omitted for single runs.

GLUE scores use at most 16 response tokens, excluding the prompt, with task-specific classification parsing. Saved answers are tokenized with the corresponding model tokenizer and truncated before scoring. SD uses the same model-size convention as Table~\ref{tab:main_results}. We average two runs for Llama-3-8B Base/MRPC and Dual-MTP/CoLA and three for all other cells.

\subsection{Ablation Study: Necessity of the Dual Training Objective}
\label{app:ablation-mask-only}

We train Mask-only with the NTP and full-context head-loss weights set to zero. Mask-only retains the masking and training settings of the corresponding Dual-MTP run. MTP and Dual-MTP reference accuracies come from Table~\ref{tab:main_results}. The 1.5B/MATH500 MTP reference has a different full-loss weight from Dual-MTP; the 7B/GSM8K MTP reference was trained on MATH7500, whereas Dual-MTP and Mask-only used GSM8K.

Compared with Dual-MTP, Mask-only loses $71.14$ points on 7B/GSM8K, $38.60$ on 7B/MATH500, $48.50$ on 1.5B/GSM8K, and $20.40$ on 1.5B/MATH500. The near-zero final-answer accuracy shows that masked-context supervision alone is insufficient when jointly training the backbone and heads.

\begin{table}[h]
\centering\small
\caption{Training-objective comparison. Accuracy (\%) is mean $\pm$ SD over three decoding runs; 1.5B uses sample SD and 7B uses population SD. MTP and Dual-MTP use the references in Table~\ref{tab:main_results}.}
\label{tab:ablation-mask-only}
\begin{tabular}{llcc}
\toprule
Model & Method & GSM8K & MATH500 \\
\midrule
Qwen2.5-7B & Mask-only & 0.10{\scriptsize $\pm$0.04} & 0.07{\scriptsize $\pm$0.09} \\
 & MTP & 72.25{\scriptsize $\pm$3.43} & 37.53{\scriptsize $\pm$1.37} \\
 & Dual-MTP & 71.24{\scriptsize $\pm$1.18} & 38.67{\scriptsize $\pm$0.90} \\
\midrule
Qwen2.5-1.5B & Mask-only & 0.23{\scriptsize $\pm$0.08} & 0.00{\scriptsize $\pm$0.00} \\
 & MTP & 47.64{\scriptsize $\pm$1.54} & 20.80{\scriptsize $\pm$1.64} \\
 & Dual-MTP & 48.72{\scriptsize $\pm$1.38} & 20.40{\scriptsize $\pm$0.87} \\
\bottomrule
\end{tabular}
\end{table}

\subsection{Draft-Head Count Sweep}
\label{app:ablation-h-sweep}

We vary the number of draft heads from one to five on both backbones. At $D=3$, we reuse the checkpoints from the main comparison. The 7B MTP checkpoint was trained on MATH7500, while the other head-count configurations use GSM8K.

\begin{table}[h]
\centering\small
\caption{Draft-head count sweep on all 1,319 GSM8K questions. Accuracy and acceptance rate (AR), in \%, are mean $\pm$ sample SD over three decoding runs. Bold marks the more accurate method.}
\label{tab:ablation-h-sweep}
\begin{tabular}{lc|cc|cc}
\toprule
Backbone & $D$ & MTP Acc. & MTP AR & Dual-MTP Acc. & Dual-MTP AR \\
\midrule
Qwen2.5-7B & 1 & 70.91{\scriptsize $\pm$0.70} & 64.56{\scriptsize $\pm$0.75} & \textbf{71.97{\scriptsize $\pm$2.39}} & 66.81{\scriptsize $\pm$0.98} \\
 & 2 & \textbf{71.34{\scriptsize $\pm$1.39}} & 57.72{\scriptsize $\pm$0.69} & 70.26{\scriptsize $\pm$1.48} & 60.04{\scriptsize $\pm$0.67} \\
 & 3 & \textbf{72.25{\scriptsize $\pm$4.20}} & 40.96{\scriptsize $\pm$0.90} & 71.24{\scriptsize $\pm$1.44} & 56.73{\scriptsize $\pm$0.56} \\
 & 4 & \textbf{71.32{\scriptsize $\pm$0.68}} & 53.03{\scriptsize $\pm$0.49} & 70.86{\scriptsize $\pm$1.14} & 55.62{\scriptsize $\pm$0.63} \\
 & 5 & \textbf{72.00{\scriptsize $\pm$1.58}} & 52.72{\scriptsize $\pm$0.34} & 70.18{\scriptsize $\pm$0.80} & 55.42{\scriptsize $\pm$0.42} \\
\midrule
Qwen2.5-1.5B & 1 & \textbf{47.38{\scriptsize $\pm$0.55}} & 56.55{\scriptsize $\pm$0.93} & 46.63{\scriptsize $\pm$1.05} & 58.52{\scriptsize $\pm$0.93} \\
 & 2 & \textbf{48.29{\scriptsize $\pm$0.67}} & 49.58{\scriptsize $\pm$0.56} & 47.23{\scriptsize $\pm$0.80} & 51.46{\scriptsize $\pm$0.56} \\
 & 3 & 47.64{\scriptsize $\pm$1.54} & 50.00{\scriptsize $\pm$0.52} & \textbf{48.72{\scriptsize $\pm$1.38}} & 49.92{\scriptsize $\pm$0.44} \\
 & 4 & 47.92{\scriptsize $\pm$1.90} & 45.41{\scriptsize $\pm$0.28} & \textbf{48.12{\scriptsize $\pm$0.91}} & 47.87{\scriptsize $\pm$0.24} \\
 & 5 & \textbf{49.58{\scriptsize $\pm$0.77}} & 45.46{\scriptsize $\pm$0.22} & 48.09{\scriptsize $\pm$0.90} & 47.80{\scriptsize $\pm$0.24} \\
\bottomrule
\end{tabular}
\end{table}

Three observations follow from \cref{tab:ablation-h-sweep}. (i) Dual-MTP improves accuracy in three of ten pairs: 1.5B at $D=3,4$ and 7B at $D=1$. Differences range from $-1.82$ to $+1.09$ points. (ii) Acceptance rate improves in nine pairs, by $3.42$ points on average; the range is $-0.08$ to $+15.77$ points. The largest increase is at $D=3$ on 7B, where accuracy is nevertheless lower. (iii) The highest observed accuracy is at $D=5$ for 1.5B MTP, $D=3$ for 1.5B Dual-MTP and 7B MTP, and $D=1$ for 7B Dual-MTP. The preferred head count therefore depends on both the backbone and training objective.

\Needspace{7\baselineskip}
\subsection{Warm-up Coefficient \texorpdfstring{$\lambda(s)$}{lambda(s)}}
\label{app:ablation-warmup}

We compare linear warm-up of the masked-loss weight with a constant $\lambda=1$ on the full MATH500 and GSM8K test sets.

\begin{table}[h]
\centering
\caption{Linear warm-up versus a constant dual-loss coefficient $\lambda=1.0$. Values are from one evaluation per configuration. Accuracy uses the final-answer metric on at most 2048 response tokens. Throughput and acceptance rate (AR, a fraction) retain the original generation measurements.}
\label{tab:ablation-warmup}
\small
\setlength{\tabcolsep}{4pt}
\begin{tabular}{l|l|ccc}
\toprule
Backbone / Task & Setting & Acc(\%) & tok/s & AR \\
\midrule
\multirow{2}{*}{Qwen2.5-7B / MATH500}
& warmup & \textbf{40.80} & 31.0 & 0.549 \\
& $\lambda{=}1.0$ & 37.40 & 34.5 & 0.561 \\
\midrule
\multirow{2}{*}{Qwen2.5-7B / GSM8K}
& warmup & \textbf{69.98} & 34.8 & 0.572 \\
& $\lambda{=}1.0$ & 69.90 & 37.6 & \textbf{0.585} \\
\midrule
\multirow{2}{*}{Qwen2.5-1.5B / MATH500}
& warmup & \textbf{18.20} & 27.8 & 0.483 \\
& $\lambda{=}1.0$ & 17.00 & 30.2 & \textbf{0.496} \\
\midrule
\multirow{2}{*}{Qwen2.5-1.5B / GSM8K}
& warmup & \textbf{48.45} & 34.3 & 0.514 \\
& $\lambda{=}1.0$ & 46.17 & 33.6 & \textbf{0.522} \\
\bottomrule
\end{tabular}
\end{table}

Warm-up improves accuracy in all four settings. The largest gain is on Qwen2.5-7B/MATH500: $40.80\%$ versus $37.40\%$, or $3.40$ points. The 1.5B gains are $1.20$ points on MATH500 and $2.27$ on GSM8K. The 7B/GSM8K scores are nearly equal ($69.98\%$ and $69.90\%$). We adopt linear warm-up.

\begingroup{}
\section{Training Computation and Cost}\label{app:training-computation-and-cost}

The full and masked views use the same backbone and draft heads, and their weighted losses are added before backpropagation. Computing the second view adds work per example. When prefix reuse is enabled, the two views share computation up to the first masked token, and gradients from both losses flow through the shared prefix. A longer shared prefix therefore saves more computation.

Across five model families and two training corpora, one epoch of Dual-MTP takes $1.26\times$--$1.50\times$ as long as one epoch of MTP (\cref{tab:training-cost}).

\begin{table}[htbp]
\centering\small
\caption{One-epoch GPU-hours (MTP / Dual-MTP) and wall-clock time ratios (Dual / MTP).}
\label{tab:training-cost}
\begin{tabular}{llcc}
\toprule
Model & Training corpus & Dual / MTP & GPU-hours (MTP / Dual)\\
\midrule
Qwen2.5-1.5B-Instruct & GSM8K & $1.26$ & 0.329 / 0.414\\
Qwen2.5-1.5B-Instruct & MATH7500 & $1.37$ & 0.342 / 0.468\\
Qwen2.5-7B-Instruct & GSM8K & $1.30$ & 0.344 / 0.447\\
Qwen2.5-7B-Instruct & MATH7500 & $1.50$ & 0.375 / 0.563\\
DeepSeek-R1-Distill-Qwen-7B & GSM8K & $1.27$ & 0.340 / 0.434\\
DeepSeek-R1-Distill-Qwen-7B & MATH7500 & $1.41$ & 0.377 / 0.531\\
Llama-3.2-1B-Instruct & GSM8K & $1.30$ & 0.200 / 0.260\\
Llama-3.2-1B-Instruct & MATH7500 & $1.35$ & 0.219 / 0.295\\
Llama-3-8B-Instruct & GSM8K & $1.31$ & 0.360 / 0.473\\
Llama-3-8B-Instruct & MATH7500 & $1.41$ & 0.436 / 0.613\\
\bottomrule
\end{tabular}
\end{table}

\Needspace{16\baselineskip}
\section{\texorpdfstring{Effect of the Verification Parameter $K_v$}{Effect of the Verification Parameter Kv}}\label{app:verification-parameter}

The verification parameter $K_v$ controls which draft candidates can be accepted by the verifier. We evaluate its effect on mathematical reasoning and code generation using Qwen2.5-7B-Instruct with $D=3$ draft heads. We first examine Dual-MTP across five values of $K_v$ at a fixed checkpoint, then present cross-method comparisons at $K_v\in\{1,3,5\}$.

\subsection{Fixed-Checkpoint Analysis on MATH500}

We hold the Qwen2.5-7B-Instruct checkpoint fixed after training on MATH7500 with three draft heads. We vary $K_v$ on the full MATH500 test set and score each final answer as in \cref{sec:experiments}.

\begin{table}[htbp]
\centering\small
\caption{Dual-MTP performance across verification parameters on MATH500. Results are mean $\pm$ sample SD over three decoding runs at a fixed checkpoint.}
\label{tab:fixed-checkpoint-verification}
\begin{tabular}{cccc}
\toprule
$K_v$ & Correct / 500 by run & Correctness (\%) & Throughput (tok/s)\\
\midrule
1 & 345 / 330 / 341 & 67.73{\scriptsize $\pm$1.55} & $118.45\pm0.90$\\
2 & 311 / 307 / 305 & 61.53{\scriptsize $\pm$0.61} & $124.97\pm0.99$\\
3 & 295 / 307 / 300 & 60.13{\scriptsize $\pm$1.21} & $125.91\pm0.41$\\
4 & 289 / 293 / 306 & 59.20{\scriptsize $\pm$1.78} & $127.03\pm0.47$\\
5 & 292 / 292 / 302 & 59.07{\scriptsize $\pm$1.15} & $127.63\pm0.15$\\
\bottomrule
\end{tabular}
\end{table}

Increasing $K_v$ from 1 to 5 raises throughput from 118.45 to 127.63 tok/s, while final-answer accuracy changes from 67.73\% to 59.07\%. The timing reference for this fixed-checkpoint experiment is the pretrained NTP model without LoRA, evaluated on the same GPU, dataset, scoring procedure, generation settings, and timing protocol. Its throughput is 83.9 tok/s. The corresponding speedups for $K_v=1,2,3,4,5$ are $1.41\times$, $1.49\times$, $1.50\times$, $1.51\times$, and $1.52\times$.

\subsection{Mathematical Reasoning}

\Cref{tab:math-verification-sweep} presents the MTP, Medusa, and Dual-MTP results on MATH500 and GSM8K at $K_v=1,3,5$. Accuracy uses the shared final-answer rule on at most 2048 response tokens. Speedup retains the original timing measurements. GSM8K Dual-MTP uses one fixed checkpoint across three decoding runs, summarized by mean and sample SD.

\begin{table}[htbp]
\centering\small
\caption{Mathematical reasoning at three verification settings. Cells give accuracy (\%, mean $\pm$ sample SD over three runs) and speedup over NTP.}
\label{tab:math-verification-sweep}
\setlength{\tabcolsep}{4pt}
\begin{tabular}{llccc}
\toprule
Task & Method & $K_v=1$ & $K_v=3$ & $K_v=5$\\
\midrule
MATH500 & MTP & 47.73{\scriptsize $\pm$2.44} / 1.46 & 39.13{\scriptsize $\pm$4.01} / 1.61 & 37.80{\scriptsize $\pm$3.61} / 1.61\\
& Medusa & 44.73{\scriptsize $\pm$3.35} / 1.80 & 35.33{\scriptsize $\pm$1.53} / 1.91 & 35.20{\scriptsize $\pm$0.87} / 1.92\\
& Dual-MTP & 67.73{\scriptsize $\pm$1.55} / 1.41 & 60.13{\scriptsize $\pm$1.21} / 1.50 & 59.07{\scriptsize $\pm$1.15} / 1.52\\
GSM8K & MTP & 78.52{\scriptsize $\pm$0.64} / 1.56 & 73.11{\scriptsize $\pm$0.97} / 1.65 & 72.73{\scriptsize $\pm$1.26} / 1.66\\
& Medusa & 77.56{\scriptsize $\pm$0.47} / 1.85 & 69.50{\scriptsize $\pm$2.50} / 1.96 & 69.40{\scriptsize $\pm$1.50} / 1.98\\
& Dual-MTP & 86.45{\scriptsize $\pm$1.66} / 1.43 & 82.16{\scriptsize $\pm$1.41} / 1.63 & 82.18{\scriptsize $\pm$0.65} / 1.65\\
\bottomrule
\end{tabular}
\end{table}

\Needspace{6\baselineskip}
\subsection{Code Generation}\label{app:code-generation}

We train each verifier on the same 266 decontaminated MBPP examples, then freeze it and use its greedy responses to train its three draft heads. At evaluation, each model generates one solution of at most 512 tokens for each of the 164 HumanEval+ and 378 MBPP+ problems. We run the same EvalPlus tests for all methods and report mean pass@1 and sample SD across three independent training runs (\cref{tab:code-generation-results}). To measure speedup, we time the first 50 problems against fine-tuned NTP on the same GPU and average the per-run ratios. The NTP baseline scores $73.58\pm1.27$\% on HumanEval+ and $70.11\pm0.46$\% on MBPP+.

\paragraph{Accuracy--Speed Trade-Off.} Dual-MTP has the highest mean pass@1 among the three draft-head methods at every evaluated $K_v$ on both benchmarks. On HumanEval+, it also has the highest mean speedup at each $K_v$ and improves on the NTP reference in both metrics. Its $K_v=1$ point attains $76.63$\% pass@1 at $1.364\times$ speedup; $K_v=3$ trades some accuracy for $1.532\times$ speedup at $74.19$\% pass@1. On MBPP+, Dual-MTP at $K_v=1$ improves on NTP in both metrics. At $K_v=3/5$, its pass@1 falls below NTP, while MTP reaches higher speeds at lower accuracy. \Cref{fig:code_pareto} shows these measured operating points and their Pareto frontiers.

\begin{figure}[htbp]
\color{black}
\centering
\includegraphics[width=\linewidth]{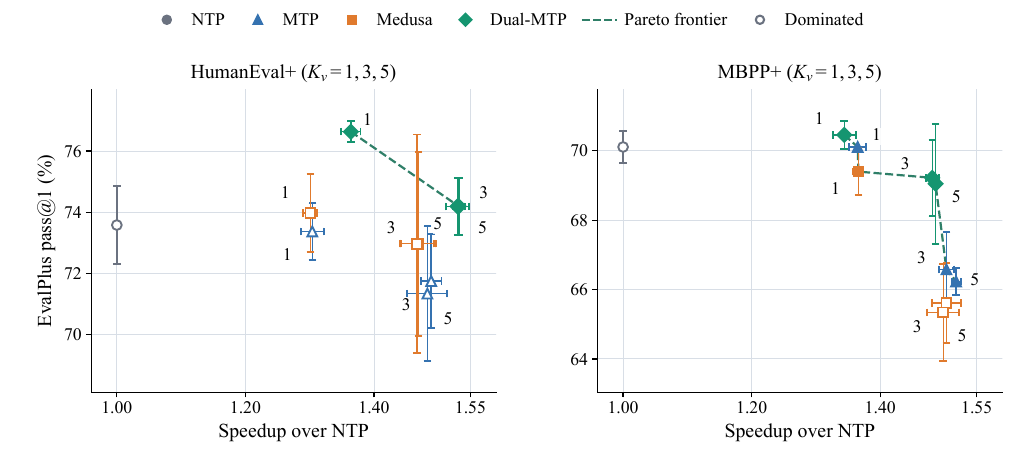}
\caption{Code-generation Pareto frontiers for Qwen2.5-7B-Instruct with the shared head-calibration procedure. The nine draft-head operating points in each panel use all three $K_v$ columns of \cref{tab:code-generation-results}; adjacent numbers indicate $K_v$. NTP is the fine-tuned reference. Vertical and horizontal bars show sample SDs of pass@1 and per-run speedup, respectively. Dashed lines connect the nondominated measured means across all methods and $K_v$ settings; hollow markers denote dominated points.}
\label{fig:code_pareto}
\end{figure}

\endgroup{}

\Needspace{6\baselineskip}
\begingroup
\section{Additional Experiments}\label{app:additional-experiments}

\subsection{Training-Duration Comparison}\label{app:epoch-comparison}

We compare two epochs of MTP with one epoch of Dual-MTP on Qwen2.5-1.5B-Instruct and Qwen2.5-7B-Instruct. Both methods train on the same examples for each task, using MATH7500 for MATH500 and the GSM8K training split for GSM8K.

The two 1.5B configurations and the 7B/MATH500 configuration were selected after examining target-task evaluation results; the 7B/GSM8K configuration was fixed.

Each entry in \cref{tab:epoch-comparison} is a single evaluation on the full test set using the final-answer metric in \cref{sec:experiments}. Dual-MTP obtains higher accuracy in three comparisons, by 0.53--3.00 points; on 7B/MATH500 it scores 41.00\% versus MTP\textquotesingle s 41.40\%. With half as many optimizer updates, Dual-MTP solves 7 more GSM8K problems at 1.5B and 12 more at 7B. Epoch count specifies data exposure; the per-epoch wall-clock costs are reported separately in \cref{tab:training-cost}.

\begin{table}[H]
\centering\small
\caption{Accuracy (\%) by training duration. Differences in percentage points use unrounded scores.}
\label{tab:epoch-comparison}
\begin{tabular}{llrrr}
\toprule
Backbone & Task & MTP (2 epochs) & Dual-MTP (1 epoch) & Difference\\
\midrule
Qwen2.5-1.5B & MATH500 & 20.80 & 23.80 & +3.00\\
Qwen2.5-1.5B & GSM8K & 48.75 & 49.28 & +0.53\\
Qwen2.5-7B & MATH500 & 41.40 & 41.00 & -0.40\\
Qwen2.5-7B & GSM8K & 72.02 & 72.93 & +0.91\\
\bottomrule
\end{tabular}
\end{table}

\Needspace{6\baselineskip}
\subsection{Evaluation on Qwen3.5}\label{app:qwen35}

We extend the mathematical-reasoning evaluation to the text decoders of Qwen3.5-4B and Qwen3.5-9B~\cite{qwen2026qwen35four,qwen2026qwen35nine}. All methods train for one epoch on the same task-specific corpora as in \cref{app:epoch-comparison}.

Dual-MTP configurations were selected using task accuracy and throughput. For 4B/GSM8K, the heads also learn from the verifier\textquotesingle s top-ranked tokens.

We report one full-test-set accuracy evaluation per configuration and measure speed on the first 32 questions of each task on an A100. Speedup is relative to fine-tuned NTP. On 9B/GSM8K, the NTP accuracy and timing results use different checkpoints.

\begin{table}[htbp]
\centering\small
\caption{Qwen3.5 mathematical-reasoning results. NTP reports accuracy (\%); the other methods report accuracy (\%) / speedup relative to the task-matched NTP timing reference. Accuracy uses the full test set, while speedup uses a fixed 32-question subset.}
\label{tab:qwen35-results}
\begin{tabular}{llrrrr}
\toprule
Backbone & Task & NTP & MTP & Medusa & Dual-MTP\\
\midrule
Qwen3.5-9B & GSM8K & 43.97 & 68.23 / 1.368 & 58.61 / 1.638 & \textbf{70.20} / 1.382\\
& MATH500 & \textbf{50.40} & 49.20 / 1.354 & 44.00 / 1.587 & \textbf{50.40} / 1.327\\
Qwen3.5-4B & GSM8K & 67.78 & 63.61 / 1.189 & 48.82 / 1.466 & \textbf{70.05} / 1.207\\
& MATH500 & 43.20 & 39.60 / 1.283 & 42.20 / 1.490 & \textbf{43.60} / 1.220\\
\bottomrule
\end{tabular}
\end{table}

Dual-MTP attains the highest accuracy in three model--task settings and ties NTP on 9B/MATH500 at 50.40\%. It exceeds MTP by 1.20--6.44 percentage points. Its measured speedup over the NTP timing reference ranges from $1.207\times$ to $1.382\times$. Medusa has the highest throughput, while Dual-MTP has the highest or tied-highest accuracy.

\subsection{Instruction Following on IFEval}\label{app:ifeval}

We evaluate Qwen2.5-7B-Instruct on IFEval~\cite{zhou2023ifeval}, using the complete set of 541 prompts and 834 verifiable instructions. Strict-prompt accuracy requires every instruction in a prompt to be satisfied; strict-instruction accuracy counts each instruction separately. Both metrics use the official evaluator.

All four methods are fine-tuned on the same 15,011 examples: 12,000 from Tulu-3 Persona Instruction-Following\footnote{\url{https://huggingface.co/datasets/allenai/tulu-3-sft-personas-instruction-following}} and 3,011 from UltraChat \texttt{train\_sft}\footnote{\url{https://huggingface.co/datasets/HuggingFaceH4/ultrachat_200k}}. We filter the training data for overlap with IFEval prompts and train for one epoch on assistant responses.

\Cref{tab:ifeval-results} reports both accuracy metrics and throughput for one full-set evaluation per method, with a limit of 2048 generated tokens. The draft-based methods use $K_v=5$, and the NTP reference is fine-tuned on the same instruction data.

\begin{table}[htbp]
\centering\small
\caption{Instruction following on IFEval with Qwen2.5-7B-Instruct. Accuracy is reported as percentage (correct/total), with throughput in tokens per second.}
\label{tab:ifeval-results}
\begin{tabular}{lrrr}
\toprule
Method & Strict prompt & Strict instruction & tok/s\\
\midrule
NTP & 70.24 (380/541) & 77.58 (647/834) & 64.12\\
MTP & 70.06 (379/541) & 77.22 (644/834) & 69.73\\
Medusa & 69.32 (375/541) & 77.46 (646/834) & \textbf{74.38}\\
Dual-MTP & \textbf{72.46 (392/541)} & \textbf{78.54 (655/834)} & 70.42\\
\bottomrule
\end{tabular}
\end{table}

Dual-MTP improves strict-prompt and strict-instruction accuracy over the fine-tuned NTP reference by 2.22 and 0.96 percentage points, respectively, with 9.82\% higher throughput. Compared with MTP, the corresponding accuracy gains are 2.40 and 1.32 points. Medusa achieves higher throughput, while Dual-MTP achieves higher accuracy on both metrics.
\endgroup

\section{Broader Impact}
\label{app:broader}
This work has the potential to improve the efficiency and reliability of large language model inference by reducing error propagation in multi-token prediction, which can lower computational costs and energy consumption in large-scale deployments. Training with perturbed contexts aims to improve generation quality while retaining the draft-and-verify inference procedure. However, improving generation quality and efficiency may also facilitate the broader deployment of language models, which could amplify existing risks such as the spread of misinformation or misuse in automated content generation. Overall, the societal impact of this work largely inherits that of the underlying language models, without introducing fundamentally new risks, but may modestly influence the balance between efficiency, accuracy, and accessibility.

\par\endgroup

\clearpage
\section*{NeurIPS Paper Checklist}

\begin{enumerate}

\Needspace{6\baselineskip}
\item {\bf Claims}
    \item[] Question: Do the main claims made in the abstract and introduction accurately reflect the paper's contributions and scope?
    \item[] Answer: \answerYes{} %
    \item[] Justification: Our claims have been reflected in both theoretical analysis and \mbox{empirical results}.
    \Needspace{3\baselineskip}

\Needspace{6\baselineskip}
\item {\bf Limitations}
    \item[] Question: Does the paper discuss the limitations of the work performed by the authors?
    \item[] Answer: \answerYes{} %
    \item[] Justification: See Section~\ref{sec:limitations}.
    \Needspace{3\baselineskip}

\Needspace{6\baselineskip}
\item {\bf Theory assumptions and proofs}
    \item[] Question: For each theoretical result, does the paper provide the full set of assumptions and a complete (and correct) proof?
    \item[] Answer: \answerYes{} %
    \item[] Justification: Assumptions and proofs are given in Appendices~\ref{app:proof-thm1} and~\ref{app:view-mi}; the predictive-entropy approximation is established in Appendix~\ref{app:empirical-cil}.
    \Needspace{3\baselineskip}
    \item[] Guidelines:
    \begin{itemize}
        \item The answer \answerNA{} means that the paper does not include theoretical results. 
        \item All the theorems, formulas, and proofs in the paper should be numbered \mbox{and cross-referenced}.
        \item All assumptions should be clearly stated or referenced in the statement of any theorems.
        \item The proofs can either appear in the main paper or the supplemental material, but if they appear in the supplemental material, the authors are encouraged to provide a short proof sketch to provide intuition. 
        \item Inversely, any informal proof provided in the core of the paper should be complemented by formal proofs provided in appendix or supplemental material.
        \item Theorems and Lemmas that the proof relies upon should be properly referenced. 
    \end{itemize}

    \Needspace{6\baselineskip}
    \item {\bf Experimental result reproducibility}
    \item[] Question: Does the paper fully disclose all the information needed to reproduce the main experimental results of the paper to the extent that it affects the main claims and/or conclusions of the paper (regardless of whether the code and data are provided or not)?
    \item[] Answer: \answerYes{} %
    \item[] Justification: Experimental settings are described in \cref{sec:experiments}. Implementation details and additional configurations are given in Appendices~\ref{app:token-masking}, \ref{app:ablation}, \ref{app:training-computation-and-cost},~\ref{app:verification-parameter}, and~\ref{app:additional-experiments}.
    \Needspace{3\baselineskip}

\Needspace{6\baselineskip}
\item {\bf Open access to data and code}
    \item[] Question: Does the paper provide open access to the data and code, with sufficient instructions to faithfully reproduce the main experimental results, as described in \mbox{supplemental material}?
    \item[] Answer: \answerYes{} %
    \item[] Justification: We have provided an open-sourced link for our code, with a well-organized \mbox{README file}.
    \Needspace{3\baselineskip}

\Needspace{6\baselineskip}
\item {\bf Experimental setting/details}
    \item[] Question: Does the paper specify all the training and test details (e.g., data splits, hyperparameters, how they were chosen, type of optimizer) necessary to understand the results?
    \item[] Answer: \answerYes{} %
    \item[] Justification: We have provided the details in \cref{sec:experiments}.
    \Needspace{3\baselineskip}
    \item[] Guidelines:
    \begin{itemize}
        \item The answer \answerNA{} means that the paper does not include experiments.
        \item The experimental setting should be presented in the core of the paper to a level of detail that is necessary to appreciate the results and make sense of them.
        \item The full details can be provided either with the code, in appendix, or as \mbox{supplemental material}.
    \end{itemize}

\Needspace{6\baselineskip}
\item {\bf Experiment statistical significance}
    \item[] Question: Does the paper report error bars suitably and correctly defined or other appropriate information about the statistical significance of the experiments?
    \item[] Answer: \answerYes{} %
    \item[] Justification: Means and standard deviations summarize repeated evaluations in \cref{tab:main_results,tab:decoding_speedup} and Appendices~\ref{app:ablation} and~\ref{app:verification-parameter}. Table~\ref{tab:ablation-warmup} and Appendix~\ref{app:additional-experiments} report a single evaluation per configuration.
    \Needspace{3\baselineskip}

\Needspace{6\baselineskip}
\item {\bf Experiments compute resources}
    \item[] Question: For each experiment, does the paper provide sufficient information on the computer resources (type of compute workers, memory, time of execution) needed to reproduce the experiments?
    \item[] Answer: \answerYes{} %
    \item[] Justification: Appendix~\ref{app:training-computation-and-cost} reports training costs, and Appendix~\ref{app:verification-parameter} describes how throughput and speedup are measured.
    \Needspace{3\baselineskip}
    
\Needspace{6\baselineskip}
\item {\bf Code of ethics}
    \item[] Question: Does the research conducted in the paper conform, in every respect, with the NeurIPS Code of Ethics \url{https://neurips.cc/public/EthicsGuidelines}?
    \item[] Answer: \answerYes{} %
    \item[] Justification: The work uses only public open-source models and benchmarks, involves no human subjects or sensitive data, and conforms to the NeurIPS Code of Ethics.
    \Needspace{3\baselineskip}
    \item[] Guidelines:
    \begin{itemize}
        \item The answer \answerNA{} means that the authors have not reviewed the NeurIPS \mbox{Code of Ethics}.
        \item If the authors answer \answerNo, they should explain the special circumstances that require a deviation from the Code of Ethics.
        \item The authors should make sure to preserve anonymity (e.g., if there is a special consideration due to laws or regulations in their jurisdiction).
    \end{itemize}

\Needspace{6\baselineskip}
\item {\bf Broader impacts}
    \item[] Question: Does the paper discuss both potential positive societal impacts and negative societal impacts of the work performed?
    \item[] Answer: \answerYes{} %
    \item[] Justification: We analyze the broader impacts of our work in Appendix \ref{app:broader}.
    \Needspace{3\baselineskip}
    
\Needspace{6\baselineskip}
\item {\bf Safeguards}
    \item[] Question: Does the paper describe safeguards that have been put in place for responsible release of data or models that have a high risk for misuse (e.g., pre-trained language models, image generators, or scraped datasets)?
    \item[] Answer: \answerNA{} %
    \item[] Justification: The paper releases no pre-trained models or scraped datasets and poses no high-risk safeguards concerns.
    \Needspace{3\baselineskip}
    \item[] Guidelines:
    \begin{itemize}
        \item The answer \answerNA{} means that the paper poses no such risks.
        \item Released models that have a high risk for misuse or dual-use should be released with necessary safeguards to allow for controlled use of the model, for example by requiring that users adhere to usage guidelines or restrictions to access the model or implementing safety filters. 
        \item Datasets that have been scraped from the Internet could pose safety risks. The authors should describe how they avoided releasing unsafe images.
        \item We recognize that providing effective safeguards is challenging, and many papers do not require this, but we encourage authors to take this into account and make a \mbox{best faith effort}.
    \end{itemize}

\Needspace{6\baselineskip}
\item {\bf Licenses for existing assets}
    \item[] Question: Are the creators or original owners of assets (e.g., code, data, models), used in the paper, properly credited and are the license and terms of use explicitly mentioned and properly respected?
    \item[] Answer: \answerYes{} %
    \item[] Justification: All base LLMs and benchmarks used in this work are properly cited in~\cref{sec:experiments} and used within the terms of their public releases.
    \Needspace{3\baselineskip}
    \item[] Guidelines:
    \begin{itemize}
        \item The answer \answerNA{} means that the paper does not use existing assets.
        \item The authors should cite the original paper that produced the code package or dataset.
        \item The authors should state which version of the asset is used and, if possible, \mbox{include a URL}.
        \item The name of the license (e.g., CC-BY 4.0) should be included for each asset.
        \item For scraped data from a particular source (e.g., website), the copyright and terms of service of that source should be provided.
        \item If assets are released, the license, copyright information, and terms of use in the package should be provided. For popular datasets, \url{paperswithcode.com/datasets} has curated licenses for some datasets. Their licensing guide can help determine the license of a dataset.
        \item For existing datasets that are re-packaged, both the original license and the license of the derived asset (if it has changed) should be provided.
        \item If this information is not available online, the authors are encouraged to reach out to the asset's creators.
    \end{itemize}

\Needspace{6\baselineskip}
\item {\bf New assets}
    \item[] Question: Are new assets introduced in the paper well documented and is the documentation provided alongside the assets?
    \item[] Answer: \answerYes{} %
    \item[] Justification: The README documents the Dual-MTP code, the only new asset.
    \Needspace{3\baselineskip}

\Needspace{6\baselineskip}
\item {\bf Crowdsourcing and research with human subjects}
    \item[] Question: For crowdsourcing experiments and research with human subjects, does the paper include the full text of instructions given to participants and screenshots, if applicable, as well as details about compensation (if any)? 
    \item[] Answer: \answerNA{} %
    \item[] Justification: The paper does not involve crowdsourcing or human subjects.
    \Needspace{3\baselineskip}
    \item[] Guidelines:
    \begin{itemize}
        \item The answer \answerNA{} means that the paper does not involve crowdsourcing nor research with human subjects.
        \item Including this information in the supplemental material is fine, but if the main contribution of the paper involves human subjects, then as much detail as possible should be included in the main paper. 
        \item According to the NeurIPS Code of Ethics, workers involved in data collection, curation, or other labor should be paid at least the minimum wage in the country \mbox{of the data collector}. 
    \end{itemize}

\Needspace{6\baselineskip}
\item {\bf Institutional review board (IRB) approvals or equivalent for research with \mbox{human subjects}}
    \item[] Question: Does the paper describe potential risks incurred by study participants, whether such risks were disclosed to the subjects, and whether Institutional Review Board (IRB) approvals (or an equivalent approval/review based on the requirements of your country or institution) were obtained?
    \item[] Answer: \answerNA{} %
    \item[] Justification: The paper does not involve human subjects, so IRB approval is not applicable.
    \Needspace{3\baselineskip}

\Needspace{6\baselineskip}
\item {\bf Declaration of LLM usage}
    \item[] Question: Does the paper describe the usage of LLMs if it is an important, original, or non-standard component of the core methods in this research? Note that if the LLM is used only for writing, editing, or formatting purposes and does \emph{not} impact the core methodology, scientific rigor, or originality of the research, declaration is not required.
    \item[] Answer: \answerNA{} %
    \item[] Justification: This paper only adopts LLMs for writing polish.
    \Needspace{3\baselineskip}
    \item[] Guidelines:
    \begin{itemize}
        \item The answer \answerNA{} means that the core method development in this research does not involve LLMs as any important, original, or non-standard components.
        \item Please refer to our LLM policy in the NeurIPS handbook for what should or should \mbox{not be described}.
    \end{itemize}

\end{enumerate}

\end{document}